\documentclass{article}
\usepackage{fancyhdr,natbib}
\usepackage{iclr2027_conference,times}

\usepackage{amsmath,amsfonts,bm}

\def\eqref#1{equation~\ref{#1}}

\def\1{\bm{1}}

\DeclareMathAlphabet{\mathsfit}{\encodingdefault}{\sfdefault}{m}{sl}
\SetMathAlphabet{\mathsfit}{bold}{\encodingdefault}{\sfdefault}{bx}{n}

\usepackage{booktabs}
\usepackage{colortbl}
\usepackage{graphicx}
\usepackage{wrapfig}
\usepackage{needspace}
\usepackage{titletoc}
\usepackage{capt-of}
\usepackage{makecell}
\usepackage{multirow}
\usepackage[section]{placeins}
\usepackage{xspace}
\usepackage{bm}
\usepackage{pifont}
\usepackage{tikz}
\usepackage{pgfplots}
\pgfplotsset{compat=1.18}
\usetikzlibrary{arrows.meta,backgrounds,calc,fit,positioning,shapes.geometric}
\usepackage{hyperref}
\usepackage{url}

\definecolor{egoblue}{HTML}{0072B2}
\definecolor{egogreen}{HTML}{009E73}
\definecolor{egoorange}{HTML}{E69F00}
\definecolor{egopurple}{HTML}{7A5195}
\definecolor{referencepurple}{HTML}{513ECB}
\newcommand{\transferinline}[1]{%
  \begin{tikzpicture}[x=1pt,y=1pt,baseline=-3pt,line cap=round,line join=round]
    \node[inner sep=0pt] at (0,0) {\includegraphics[viewport=120 0 710 1010,clip,height=13pt]{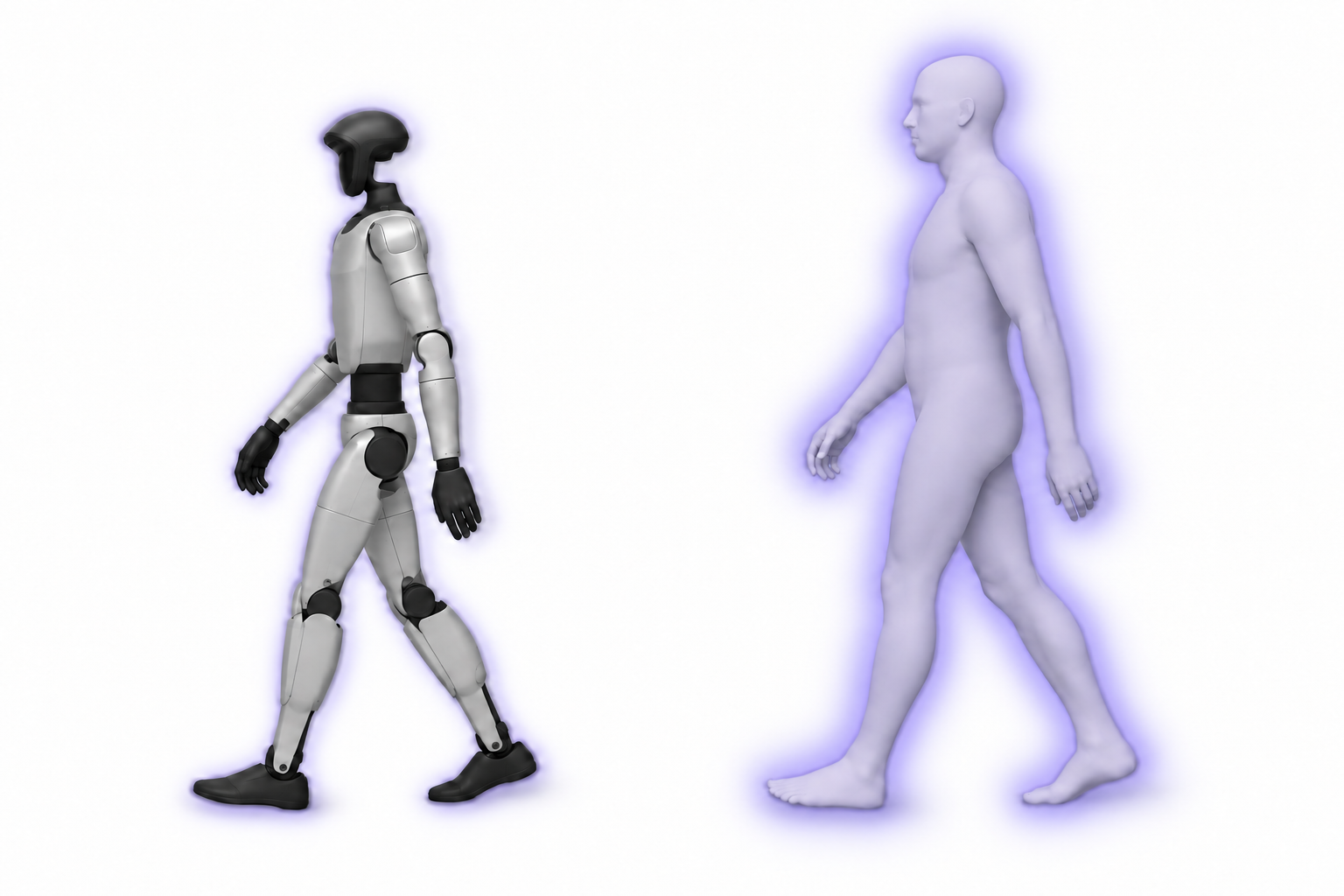}};
    \node[inner sep=0pt,opacity=.5] at (0,0) {\includegraphics[viewport=120 0 710 1010,clip,height=13pt]{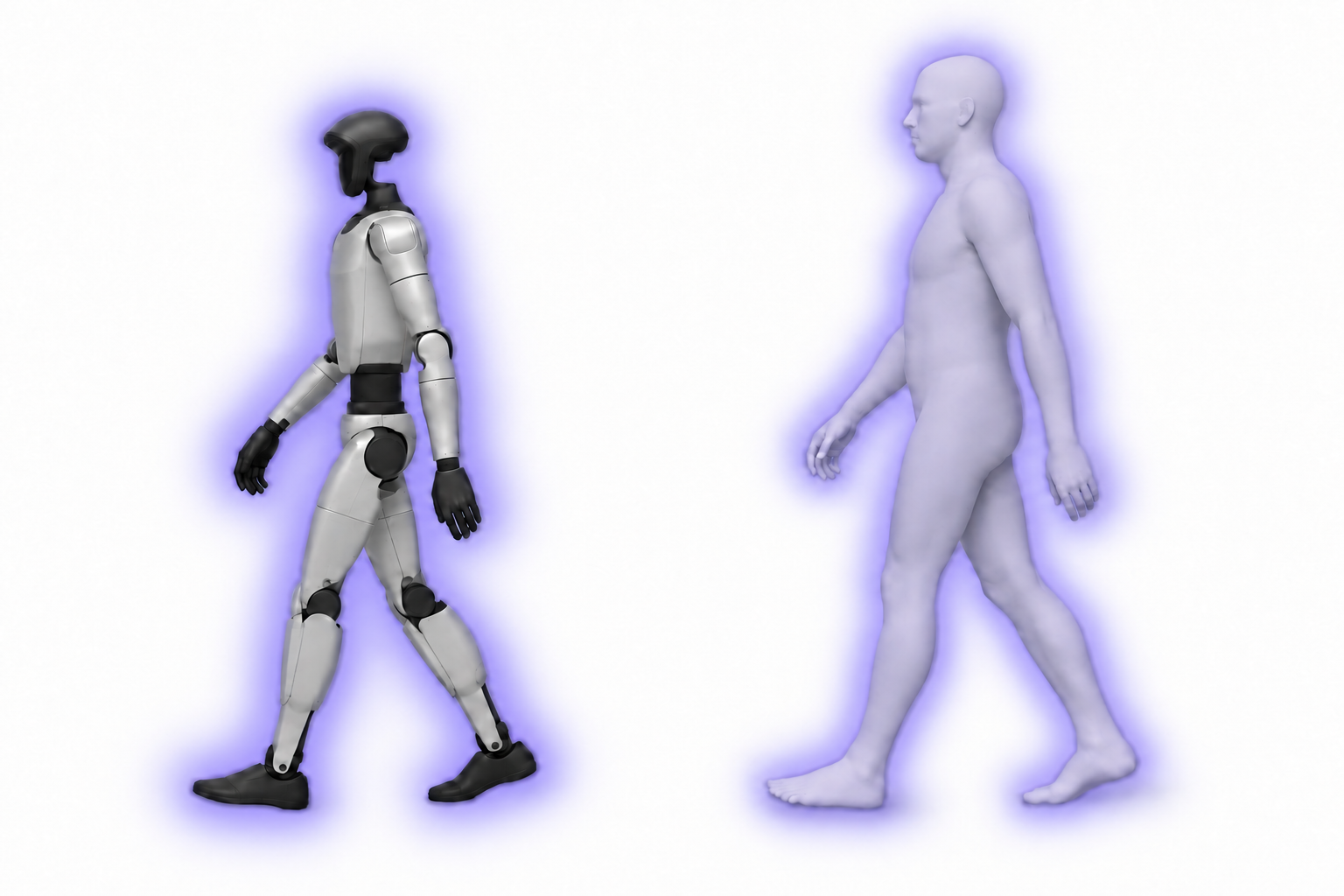}};
    \ifnum#1<3
      \draw[gray!65,line width=.3pt] (4.6,0) -- (6.6,0) (5.6,-1) -- (5.6,1);
      \node[inner sep=0pt] at (11,0) {\includegraphics[viewport=800 0 1380 1024,clip,height=12.2pt]{figs/generalization_strong_glow_icons.png}};
      \def\inlineobjectx{23}
      \draw[-{Stealth[length=1.5pt,width=1.3pt]},gray!65,line width=.3pt] (15.5,0) -- (19,0);
    \else
      \def\inlineobjectx{12}
      \draw[-{Stealth[length=1.5pt,width=1.3pt]},gray!65,line width=.3pt] (4.5,0) -- (8,0);
    \fi
    \begin{scope}[shift={(\inlineobjectx,0)},draw=referencepurple,line width=.4pt]
      \ifnum#1=1
        \path[draw,fill=referencepurple!12] (0,3) -- (2.5,1.5) -- (2.5,-1.5) -- (0,-3) -- (-2.5,-1.5) -- (-2.5,1.5) -- cycle;
        \draw (-2.5,1.5) -- (0,0) -- (2.5,1.5) (0,0) -- (0,-3);
        \fill[white] (2,-2) circle (1.5pt);
        \draw[line width=.5pt] (1,-2) -- (1.8,-2.8) -- (3.2,-1.1);
      \else
        \draw[dash pattern=on .8pt off .6pt] (0,3) -- (2.5,1.5) -- (2.5,-1.5) -- (0,-3) -- (-2.5,-1.5) -- (-2.5,1.5) -- cycle;
        \draw[referencepurple!50] (-2.5,1.5) -- (0,0) -- (2.5,1.5) (0,0) -- (0,-3);
      \fi
    \end{scope}
  \end{tikzpicture}%
}

\hypersetup{
  colorlinks=true,
  linkcolor=referencepurple,
  citecolor=referencepurple,
  urlcolor=referencepurple
}
\newif\ifshowrevisions
\showrevisionsfalse

\newcommand{\datasetname}{\textsc{HumanVerse-500}\xspace}
\newcommand{\modelname}{\ensuremath{\lambda_{0}}\xspace}

\title{\vspace{-2em}Towards a General Humanoid Loco-Manipulation Model
via Egocentric Whole-Body Human Data Pretraining}

\author{%
\parbox[t]{\dimexpr\textwidth-2\tabcolsep\relax}{\raggedright
\textbf{Chongyang~Xu\textsuperscript{1,2}\thanks{This work was done during Chongyang Xu's internship at Alibaba Token Hub.}, Zhao~Wu\textsuperscript{1},
Jin~Chen\textsuperscript{1,3}, Yiming~Jiang\textsuperscript{1,4},
Jinhui~Ye\textsuperscript{5}, Yuming~Jiang\textsuperscript{1},
Shifeng~Zhang\textsuperscript{1}, Ziliang~Feng\textsuperscript{2},
Mu~Xu\textsuperscript{1}, Yilun~Chen\textsuperscript{1\textdagger\ding{41}},
Li~Lu\textsuperscript{2\ding{41}}, Steven~C.H.~Hoi\textsuperscript{1}}\\[0.5em]
\normalfont\textsuperscript{1}Alibaba Group\qquad
\textsuperscript{2}Sichuan University\qquad
\textsuperscript{3}Shanghai Innovation Institute\\[0.2em]
\textsuperscript{4}Beihang University\qquad
\textsuperscript{5}Hong Kong University of Science and Technology\\[0.3em]
{\small\textsuperscript{\ding{41}}Corresponding authors.\qquad
\textsuperscript{\textdagger}Project leader.}
}
}

\iclrfinalcopy

\begin{document}

\setcounter{footnote}{0}
\maketitle
\lhead{Preprint}

\vspace{-3em}
\begin{figure}[!h]
  \centering
  \includegraphics[width=0.85\linewidth]{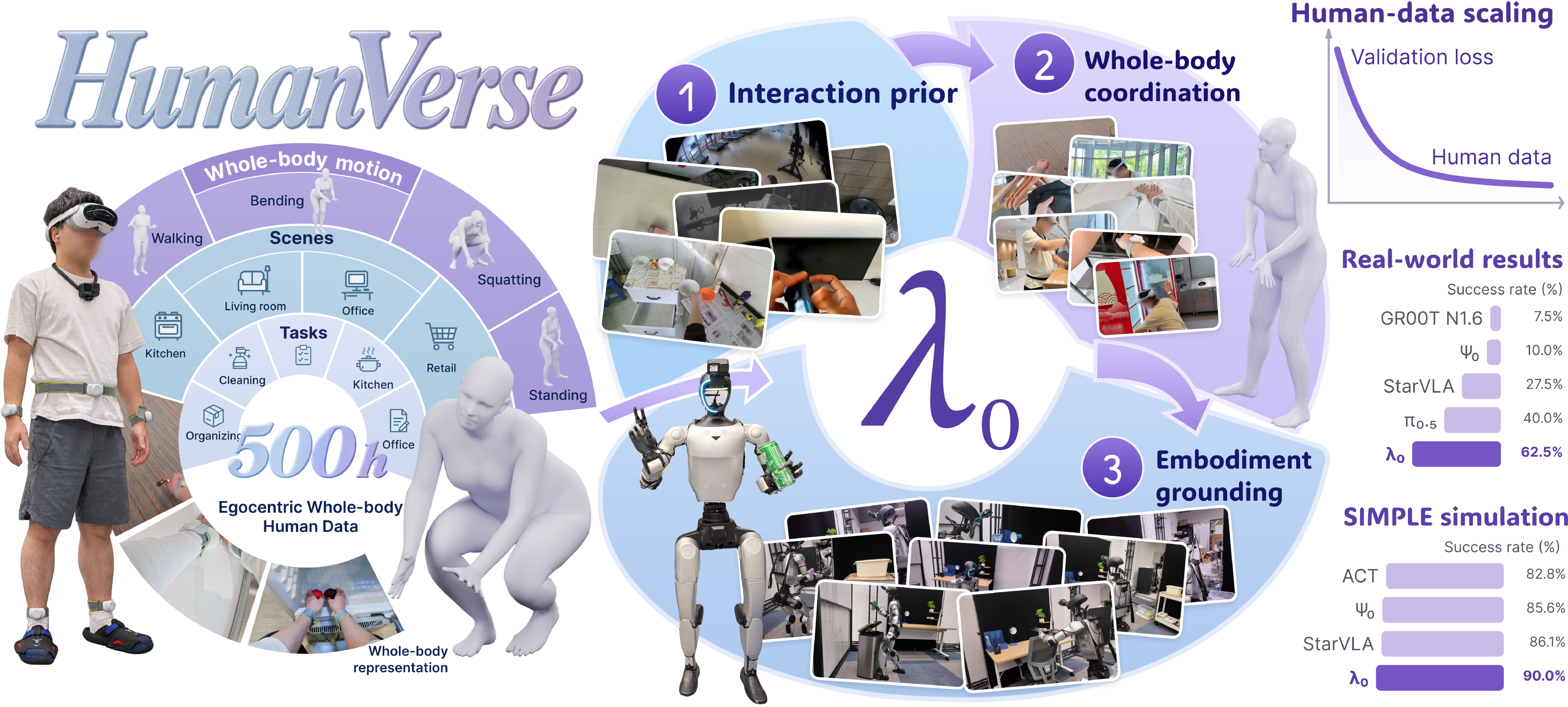}
  \caption{Introducing $\bm{\lambda_{0}}$ with \textbf{\textsc{HumanVerse-500}}.
  We introduce a robot-free egocentric human-data collection paradigm and use it to collect \datasetname, a 500-hour dataset of whole-body human loco-manipulation.
  Then, we develop $\lambda_{0}$, a unified whole-body humanoid vision--language--action policy, through a three-stage recipe combining large-scale Internet manipulation data, \datasetname, and embodiment-specific data.
  Finally, $\lambda_{0}$ achieves state-of-the-art performance and strong generalization across simulated and real-world loco-manipulation tasks.}
  \label{fig:teaser}
\end{figure}

\begin{abstract}
Humanoid whole-body manipulation has advanced rapidly, enabling policies to coordinate locomotion, posture, bimanual interaction, and dexterous hand movements.
Meanwhile, egocentric human videos provide diverse examples of everyday interactions across objects and scenes, offering scalable supervision without robot operation.
However, existing supervision from these videos provides limited coverage of whole-body movement and coordination with hand--object interaction, while obtaining such supervision through humanoid teleoperation is also costly and difficult to scale.
We therefore explore how human experience can support scalable learning of humanoid loco-manipulation.
To support this study, we introduce \datasetname, a 500-hour dataset of diverse human loco-manipulation behaviors in open-world environments, collected with a lightweight wearable system that synchronizes egocentric video with body and hand motion.
Building on this dataset, we develop \modelname, a whole-body humanoid vision--language--action policy, through three-stage training that first learns interaction from diverse egocentric datasets, then coordinates body and hand motion using \datasetname, and finally adapts the policy to downstream tasks and robot embodiments.
Across these stages, \modelname learns a shared representation space for human experience transfer, while domain-specific interfaces handle differences between human and robot states and actions.
We evaluate \modelname on SIMPLE and 4 real-world loco-manipulation tasks, achieving state-of-the-art performance, and further analyze its scaling behavior, generalization, and training-stage contributions to understand how human data support downstream whole-body humanoid control.
We will release our code, models, and data to support further research.

\end{abstract}

\section{Introduction}
Humanoid whole-body manipulation has advanced rapidly, enabling robots to coordinate locomotion, posture, and dexterous manipulation within a unified policy~\citep{jiang2026wholebodyvla,shi2026egohumanoid,wei2026psi0,hu2026openhlm,figure2026helix02,figure2026helix25}.
Learning these coordinated behaviors requires demonstrations that capture both whole-body movement and interaction with objects.
Such demonstrations can be collected through whole-body teleoperation, where human operators guide humanoids through coordinated movements and object interactions~\citep{fu2025humanplus,he2025omnih2o,luo2025sonic,gear2025grootn16,li2026omega0}.
Collecting diverse demonstrations this way requires substantial robot time and human effort, making it difficult to scale across tasks and environments~\citep{shi2026egohumanoid,punamiya2026egoverse}.

Egocentric human videos offer a scalable alternative, capturing diverse everyday interactions without requiring robot operation~\citep{grauman2022ego4d,kareer2024egomimic,hoque2025egodex,punamiya2026egoverse,figure2026index}.
Recent methods recover wrist and hand trajectories from these videos, turning observed interactions into action supervision for robot learning~\citep{li2025vitra,yang2025egovla,wang2026ego2robot}.
Pretraining on these action-labeled videos can improve downstream manipulation, supporting the use of human experience for scalable robot learning~\citep{zheng2026egoscale,punamiya2026egoverse}.
However, wrist and hand trajectories alone do not fully capture the whole-body coordination required for humanoid loco-manipulation.
A humanoid must approach an object, adjust its posture to reach it, and maintain its grasp while carrying it elsewhere~\citep{shi2026egohumanoid,zhao2026halomi}.
Learning these behaviors requires supervision that preserves the temporal coordination between body movement and hand--object interaction.
We therefore ask how diverse human experience can provide scalable whole-body supervision for humanoid loco-manipulation.

To address this question, we introduce \datasetname, a 500-hour dataset of human loco-manipulation collected across diverse activities in open-world environments.
We use a lightweight wearable system to capture egocentric video and body motion, then reconstruct hand motion from video to provide synchronized whole-body supervision.
Building on \datasetname, we develop \modelname, a whole-body humanoid vision--language--action policy that transfers human experience to robot control (Fig.~\ref{fig:teaser}).
We train the policy in three stages to learn interaction from diverse egocentric datasets, coordinate body and hand motion using \datasetname, and adapt to downstream tasks and robot embodiments.
A shared representation space enables transfer across stages, while domain-specific interfaces handle differences between human and robot states and actions.

We evaluate \modelname on SIMPLE and four real-world loco-manipulation tasks, improving success rates over the strongest evaluated baselines by 3.9 and 22.5 percentage points, respectively.
Across the evaluated data scales, more whole-body human data reduces validation loss and improves real-world task progress.
Ablations further identify whole-body mid-training as the larger contributor to real-world success among the two human-data stages, while human demonstrations improve progress on robot-unseen objects.
These results support scalable whole-body human data as a source of supervision for humanoid loco-manipulation.
Our contributions are summarised as follows:
\begin{itemize}
  \item We introduce \datasetname, a 500-hour dataset of open-world human loco-manipulation with synchronized egocentric video, body motion, and hand motion.
  \item We develop \modelname, a whole-body humanoid vision--language--action policy that transfers human experience through three-stage training in a shared human--robot representation space.
  \item We reveal a consistent scaling trend in whole-body human pretraining and demonstrate its benefits for downstream performance and generalization in humanoid loco-manipulation.
\end{itemize}

\section{Related Work}
\subsection{Generalist Robot Policies}
Large-scale robot datasets bring together demonstrations across tasks and embodiments, supporting the development of generalist policies~\citep{dasari2020robonet,jang2022bcz,walke2023bridgedata,openx2024,brohan2023rt1,zitkovich2023rt2,khazatsky2024droid,bu2025agibot,zhao2025humanoideveryday}.
These policies learn from diverse demonstrations within a shared model~\citep{ghosh2024octo,kim2025openvla,chi2025diffusion}, motivating more scalable ways to collect training data.
Advances in teleoperation and portable demonstration interfaces reduce collection effort~\citep{dass2023pato,qin2023anyteleop,zhao2023aloha,fu2025mobilealoha,chi2024umi}.
However, robot-based collection remains tied to hardware availability and repeated environment resets, while portable interfaces capture motion within their supported action spaces.

\subsection{Learning from Human Videos}
Human video datasets capture diverse activities, objects, and environments, providing rich experience for robot learning~\citep{grauman2022ego4d,damen2022epic,goyal2017something,grauman2024egoexo4d,wang2023holoassist,detone2026nymeriaplus,lightwheel2026egosuite}.
Learning from it requires connecting observations to actions.
Temporal and geometric annotations describe when interactions occur and how hands and objects move~\citep{liu2022hoi4d,wang2025hocap,sener2022assembly101,ragusa2021meccano,huang2024egoexolearn,banerjee2025hot3d}, while motion tracking and reconstruction provide human trajectories for action learning~\citep{hoque2025egodex,punamiya2026egoverse,li2025vitra,zheng2026egoscale,wang2026ego2robot}.
Complementary approaches learn visual representations from human videos~\citep{nair2023r3m,sontakke2023roboclip,kumar2020subroutines} or connect human demonstrations to robot control through imitation and alignment~\citep{mandlekar2022offline,shao2020concept2robot,bahl2022whirl,wang2023mimicplay,bharadhwaj2023zeroshot,kareer2024egomimic,lepert2025phantom,yang2025egovla}.
Human videos offer scalable supervision for manipulation, but capturing whole-body coordination remains challenging for humanoid learning.

\subsection{Humanoid Whole-Body Control}
Humanoid whole-body control coordinates locomotion, posture, and manipulation to support complex physical interactions~\citep{fu2025humanplus,he2025omnih2o,luo2025sonic}.
Building on this control foundation, humanoid policies learn task-level behaviors from human experience through latent video actions, geometric pretraining, or joint training on human and robot demonstrations~\citep{jiang2026wholebodyvla,wei2026psi0,shi2026egohumanoid,hu2026openhlm,bjorck2025groot,beingbeyond2026beingh}.
Predictive models and integrated whole-body systems offer complementary ways to connect perception, motion, and control~\citep{li2026omega0,symbiosis2026dpc,figure2026helix02,parada2026geminirobotics2,zheng2026motionwam,ye2026dreamzero,nvidia2026dreamdojo}.
A remaining challenge is to turn diverse human experience into scalable supervision for humanoid whole-body control.

\section{HumanVerse-500: Egocentric Whole-Body Human Data}
\label{sec:humanverse-data}
\begin{figure}[!b]
  \centering
  \includegraphics[width=\linewidth]{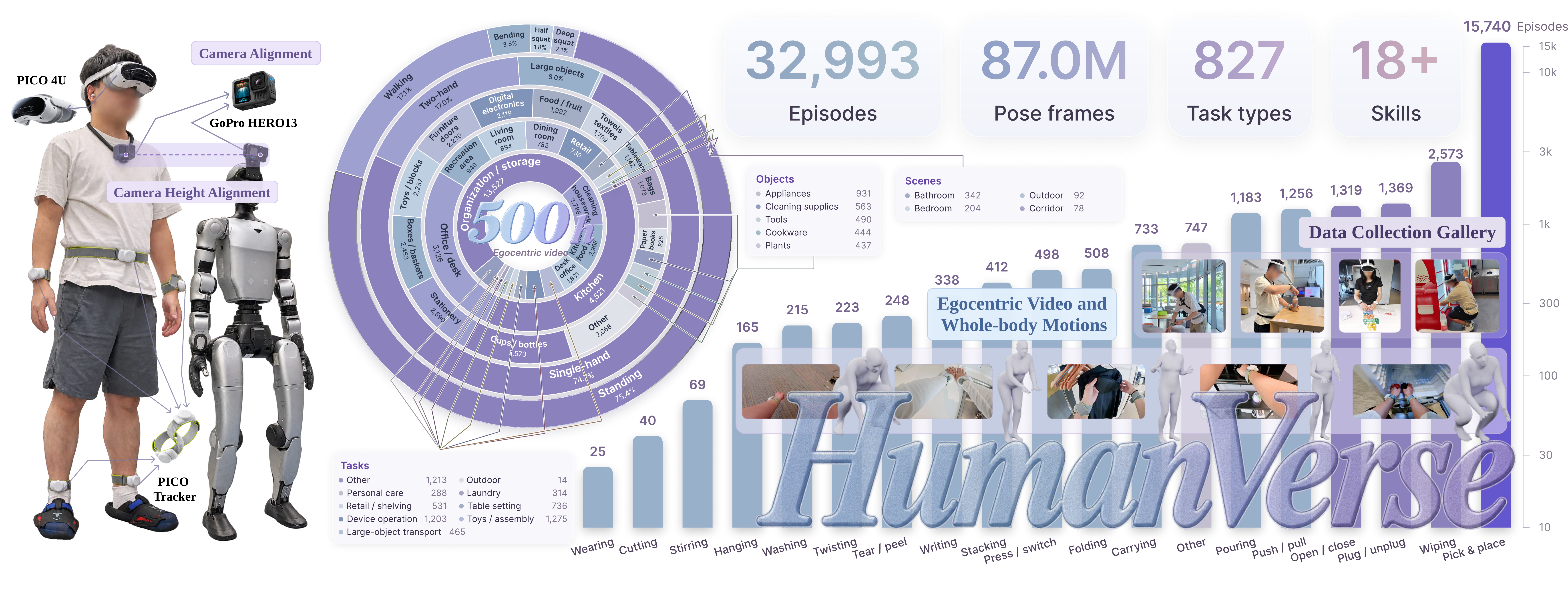}
  \caption{\textbf{\datasetname.}
  The wearable capture setup, task and motion distributions, and paired egocentric video and whole-body motions illustrate the dataset's scale and diversity.}
  \label{fig:hardware-setup}
\end{figure}

As shown in Fig.~\ref{fig:hardware-setup}, \datasetname contains 500 hours of egocentric human loco-manipulation, comprising 32,993 episodes and 87 million pose frames.
The dataset covers 827 task types across diverse objects and scenes, pairing first-person video with synchronized body and hand motion.
We choose the GoPro HERO13 for convenient human data collection and equip our downstream robot, Unitree G1, with the same camera on its head.
We align the human camera height with the robot camera using a neck mount, while a PICO 4 Ultra headset and body trackers capture body motion alongside the video.
We then reconstruct hand motion from the same video.

\subsection{Data Coverage}

\begin{figure}[!ht]
  \centering
  \includegraphics[width=\linewidth]{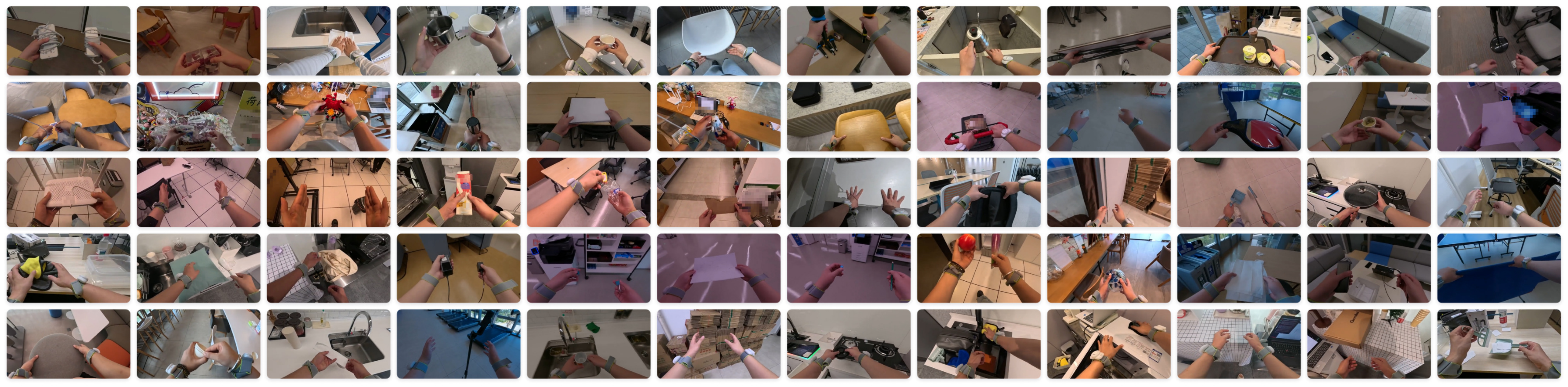}
  \caption{\textbf{Egocentric task diversity in \datasetname.}
  Sixty egocentric frames span carrying, cleaning, plant care, office tasks, and object handling.}
  \label{fig:data-diversity-examples}
\end{figure}

\datasetname captures everyday activities across indoor and outdoor environments, including cleaning, cooking, personal care, laundry, table setting, and large-object transport.
These activities involve objects of different sizes and require varied combinations of locomotion, posture changes, and hand--object interaction.
For example, interacting with objects at different heights involves bending or reaching, while transporting objects couples walking with sustained grasping.
Continuous recordings capture transitions between interactions and body--hand coordination throughout each task.
Figure~\ref{fig:hardware-setup} summarizes the task, scene, object, and motion distributions.
Figure~\ref{fig:data-diversity-examples} shows representative egocentric frames spanning household, office, and object-handling activities.

\FloatBarrier

\subsection{Whole-Body Motion Reconstruction}

\begin{figure}[!ht]
  \centering
  \includegraphics[width=\linewidth]{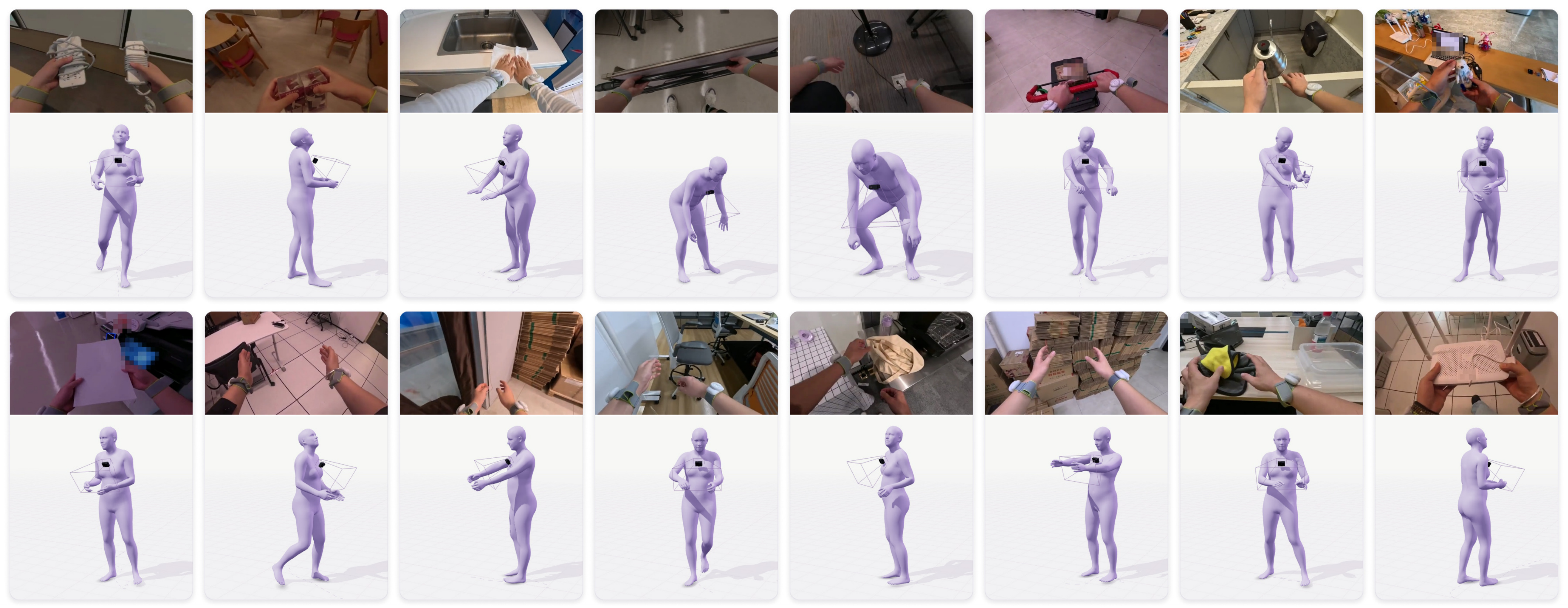}
  \caption{\textbf{Whole-body Human Data.} Egocentric RGB frames, body motion, and hand motion.}
  \label{fig:smpl-camera-examples}
\end{figure}

We construct whole-body motion by combining wearable-tracked body kinematics with hand motion reconstructed from video.
The PICO system~\citep{pico_body_tracking} provides timestamped 24-joint body motion, while Dyn-HaMR~\citep{yu2025dynhamr} reconstructs 21 joints per hand using camera motion estimated by ViPE~\citep{huang2025vipe}.
Timestamps and coordinate transforms align body and hand trajectories with video frames, while an episode-fixed frame preserves hand displacement.
Validity masks retain usable body supervision when hands are missing or occluded.
Quality checks cover timestamp consistency, finite values, hand reprojections, and synchronized motion overlays.
Figure~\ref{fig:smpl-camera-examples} shows synchronized body motion, hand motion, and egocentric RGB frames.

\FloatBarrier

\Needspace{17\baselineskip}
\subsection{Human-to-Robot Retargeting}

\begingroup
\setlength{\intextsep}{0pt}
\begin{wrapfigure}{r}{0.4\textwidth}
  \vspace{-1em}
  \setlength{\abovecaptionskip}{3pt}
  \setlength{\belowcaptionskip}{0pt}
  \setlength{\parskip}{0pt}
  \centering
  \includegraphics[width=\linewidth]{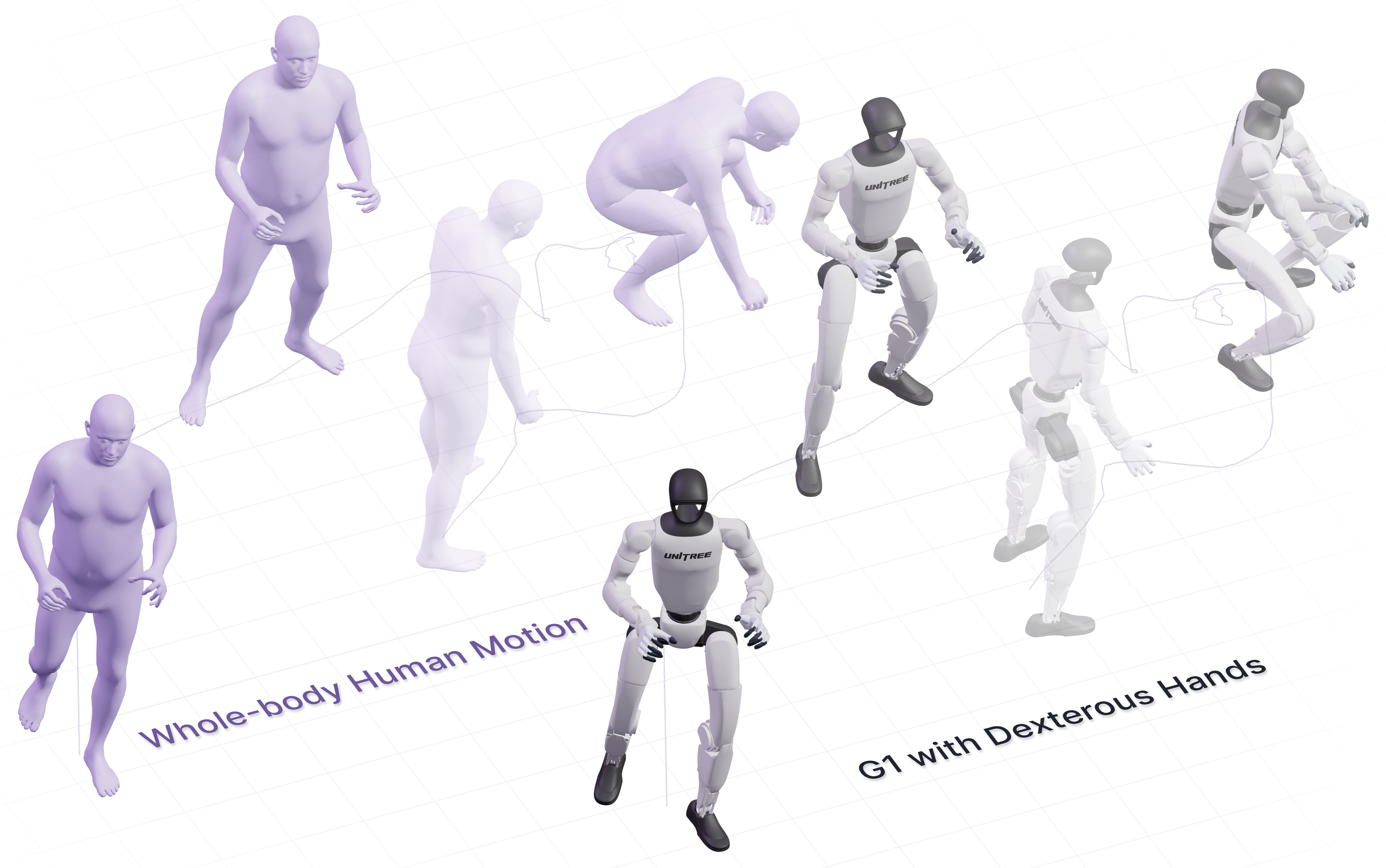}
  \caption{Whole-body motion retargeting to G1 with dexterous hands.}
  \label{fig:human-g1-motion}
\end{wrapfigure}

We convert the captured human motion into training data for the Unitree G1 policy using SONIC~\citep{luo2025sonic}.
Its frozen encoder produces body-motion tokens, and its controller tracks human motion in MuJoCo to obtain G1 states.
We retarget the reconstructed hand motion to the bilateral BrainCo Revo 2 dexterous hands~\citep{brainco_revo2} to obtain robot hand commands.
The paired records share egocentric video and timestamps while retaining distinct body and hand representations.
Figure~\ref{fig:human-g1-motion} shows the corresponding retargeting.
\par
\endgroup

\section{\texorpdfstring{$\lambda_0$}{lambda0}: Learning Humanoid Control from Human Experience}
\label{sec:method}
\begin{figure*}[t]
  \centering
  \includegraphics[width=\textwidth]{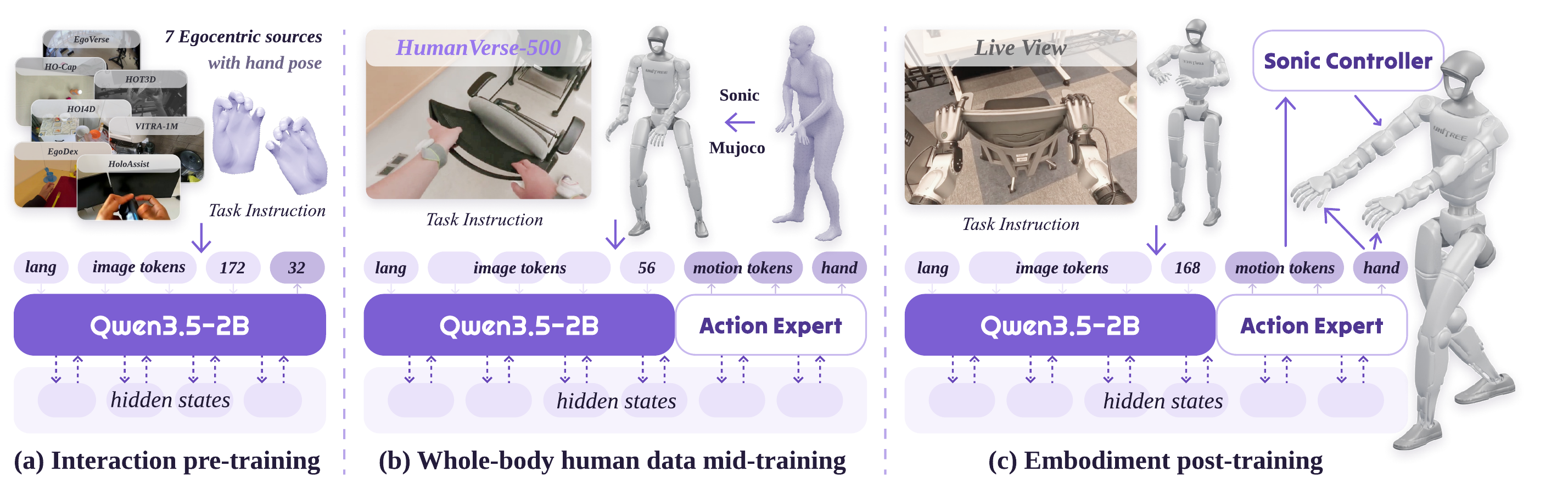}
  \caption{\textbf{Three-stage training of \modelname.}
  The policy learns interaction from egocentric data, whole-body coordination from \datasetname, and downstream control from real-robot data.}
  \label{fig:training-recipe}
\end{figure*}

We develop \modelname with a shared vision--language--action architecture and domain-specific interfaces to transfer human experience to humanoid control.
Training proceeds through interaction pre-training, whole-body mid-training, and embodiment-specific post-training (Fig.~\ref{fig:training-recipe}).

\subsection{Policy Architecture and Action Interfaces}
\label{sec:state-action-interfaces}

Given an egocentric image $\mathbf{x}$, a language instruction $\ell$, and the current state $\mathbf{s}^{q}$, \modelname predicts a 50-step action chunk in the active action space $q$.
The backbone and action expert are shared across stages, while domain-specific action projections handle differences in target dimensions and semantics.

\paragraph{Shared transfer core.}
\modelname consists of a Qwen3.5-2B~\citep{qwen2026qwen35} vision--language backbone $V_\phi$ and a generative action expert $F_\psi$.
The current state is quantized into 256 bins and appended to the language instruction.
The backbone encodes this state-aware instruction together with the image $\mathbf{x}$, then supplies layer-wise features to the action expert.
For each action space $q$, an input encoder $E_q$ maps the noisy action and noise level into action tokens, while an output decoder $D_q$ maps the shared features to a velocity prediction in that action space:
\begin{equation}
  \widehat{\mathbf{v}}^{q}
  = D_q\!\left(
  F_{\psi}\!\left(V_{\phi}(\mathbf{x},[\ell;Q(\mathbf{s}^{q})]),
  E_q(\widetilde{\mathbf{a}}^{q},\boldsymbol{\sigma}),e\right)\right).
  \label{eq:shared-policy}
\end{equation}
Here $e$ is the embodiment identifier.
The backbone $V_\phi$ and action expert $F_\psi$ are shared across stages, while separate input encoders $E_q$ and output decoders $D_q$ handle the corresponding action spaces.

\paragraph{Action interfaces.}
The egocentric interface predicts hand motion conditioned on video, language, and bilateral hand states.
The robot-compatible view of mid-training data uses the same state--action format as downstream G1 post-training data, while the two stages use distinct training domains.

\subsection{Three-Stage Training}
\label{sec:training-recipe}
\label{sec:training-data}

Our training pipeline uses three data sources: public egocentric hand--object datasets ($\mathcal{D}_E$) for pretraining on egocentric videos and hand poses, \datasetname{} ($\mathcal{D}_W$) for mid-training on whole-body human motion, and robot demonstrations ($\mathcal{D}_R$) for robot post-training.

All stages use the masked velocity-field loss $\mathcal{L}^{q}$ defined in Sec.~\ref{sec:training-objective-deployment}, with different data sources and active interfaces.
Let $\boldsymbol{\theta}^{(k)}$ denote the policy parameters learned in Stage $k$.
Stages II and III initialize from the preceding EMA checkpoint while changing the supervision and active interfaces.

\paragraph{Stage I: egocentric hand-motion pretraining.}
We initialize the backbone from a vision--language checkpoint and train it with the action expert and egocentric hand interface (Fig.~\ref{fig:training-recipe}(a)).
Training uses EgoDex, EgoVerse, HO-Cap, HOI4D, HoloAssist, HOT3D, and VITRA-1M \citep{hoque2025egodex,punamiya2026egoverse,wang2025hocap,liu2022hoi4d, wang2023holoassist,banerjee2025hot3d,li2025vitra}, with hand-pose supervision from either native ground-truth annotations or pseudo-labels generated by open-source methods.
Let $\mathcal{S}_E$ denote these seven data sources. We sample datasets in proportion to the square root of their usable frame counts $N_d$ to increase the relative contribution of smaller sources:
\begin{equation}
  p_1(d)=\frac{\sqrt{N_d}}{\sum_{d'\in\mathcal{S}_E}\sqrt{N_{d'}}},
  \qquad
  \boldsymbol{\theta}^{(1)}=
  \arg\min_{\boldsymbol{\theta}}
  \mathbb{E}_{d\sim p_1,\,\tau\sim\mathcal{D}_d}
  \left[\mathcal{L}^{E}(\tau;\boldsymbol{\theta})\right].
  \label{eq:stage1-main}
\end{equation}
Stage I predicts a 138-dimensional bimanual action at each time step, with each hand represented by 3D relative wrist translation, 6D relative wrist rotation, and $20\times3$ local hand-keypoint coordinates.

\paragraph{Stage II: whole-body human-motion mid-training.}
During mid-training, \datasetname{} demonstrations are converted into the G1 state--action format (Fig.~\ref{fig:training-recipe}(b)).
Each action comprises a 64-dimensional SONIC motion latent and two 6-dimensional Revo2 hand commands, totaling 76 dimensions.
Human-derived and native robot data use distinct domain identifiers and separate input and output projections to accommodate scale differences.
Denoting the conversion by $\Phi_R$ and the human-domain loss by $\mathcal{L}^{H}$, we optimize
\begin{equation}
  \boldsymbol{\theta}^{(2)}=
  \arg\min_{\boldsymbol{\theta}\leftarrow
  \operatorname{EMA}(\boldsymbol{\theta}^{(1)})}
  \mathbb{E}_{\tau\sim\Phi_R(\mathcal{D}_W)}
  [\mathcal{L}^{H}(\tau;\boldsymbol{\theta})].
  \label{eq:stage2-main}
\end{equation}
We update the backbone, action expert, and human-data projections while keeping the Stage-I hand head frozen.
This stage uses only human data. Robot demonstrations are introduced in Stage III.

\paragraph{Stage III: robot post-training.}
We post-train the policy on robot demonstrations to adapt the motion and interaction patterns learned in the preceding stages to downstream tasks and executable robot control (Fig.~\ref{fig:training-recipe}(c)).
The post-training objective is
\begin{equation}
  \boldsymbol{\theta}^{(3)}=
  \arg\min_{\boldsymbol{\theta}\leftarrow
  \operatorname{EMA}(\boldsymbol{\theta}^{(2)})}
  \mathbb{E}_{\tau\sim\mathcal{D}_R}
  [\mathcal{L}^{R}(\tau;\boldsymbol{\theta})].
  \label{eq:stage3-main}
\end{equation}
The shared model, embodiment conditioning, and active interfaces are optimized jointly.

\subsection{Training Objective and Deployment}
\label{sec:training-objective-deployment}

\paragraph{Masked velocity-field objective.}
We interpolate between clean actions and Gaussian noise and train the model to predict the velocity target $\mathbf{v}=\boldsymbol{\epsilon}-\mathbf{a}$.
Let $\widehat{\mathbf{v}}$ denote the predicted velocity and $M^q$ the set of valid, non-prefix sample--time--feature indices for action space $q$, excluding missing annotations.
We minimize the mean squared error over these entries:
\begin{equation}
 \mathcal{L}^{q}=\mathbb{E}\!\left[
 \frac{1}{|M^q|}\sum_{(i,h,j)\in M^q}
 \left(\widehat{v}^{q}_{i,h,j}-v^{q}_{i,h,j}\right)^2
 \right].
 \label{eq:denoising-summary}
\end{equation}
At inference, we initialize the action chunk from Gaussian noise and denoise it using 10 Euler steps from noise level one to zero. Any provided action prefix remains fixed throughout denoising.

\paragraph{Deployment.}
At deployment, \modelname{} receives G1's egocentric observation, language instruction, and proprioceptive state, using the robot-domain identifier and its corresponding input and output projections (Fig.~\ref{fig:training-recipe}(c)).
The predicted 64-dimensional motion latent is passed to the SONIC controller for whole-body execution, while the two 6-dimensional hand commands are sent to the Revo2 hands.

\section{Experiments}
\label{sec:experiments}
We evaluate whether the three-stage recipe turns human data into improved humanoid control and object transfer.
Specifically, we ask: \textbf{Q1.}
\emph{How well does the model perform on humanoid loco-manipulation?}
(Sec.~\ref{sec:exp-performance}); \textbf{Q2.}
\emph{How do human-data scale and the training recipe affect learning?}
(Sec.~\ref{sec:exp-learning}); \textbf{Q3.}
\emph{Does human-data scaling improve downstream transfer?}
(Sec.~\ref{sec:exp-transfer}).
Accordingly, we evaluate held-out learning dynamics, task performance, and unseen-object generalization.

\subsection{Evaluation Setup and Metrics}
\label{sec:evaluation-protocol}

\begin{figure}[!htbp]
  \centering
  \input{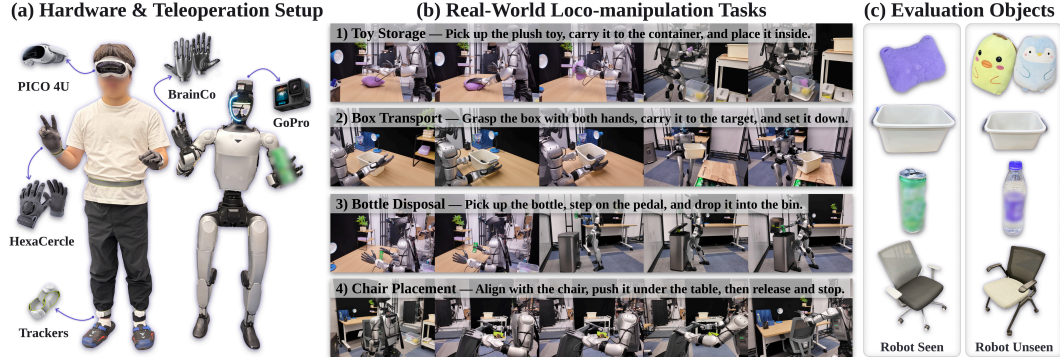}
  \setlength{\abovecaptionskip}{2pt}
  \caption{\textbf{Real-world data collection and evaluation.}
  (a) Egocentric teleoperation collects whole-body demonstrations.
  (b) Four loco-manipulation tasks evaluate task success.
  (c) Robot-seen and robot-unseen objects are used for standard task and object-generalization evaluation, respectively.}
  \label{fig:robot-embodiment-teleoperation}
  \label{fig:real-world-task-settings}
\end{figure}

\noindent\textbf{Hardware.}
As shown in Fig.~\ref{fig:real-world-task-settings}(a), a PICO 4U with five trackers captures body motion, which SONIC converts into Unitree G1 control signals.
We use HexaCercle M11 gloves to capture hand poses robustly under occlusion and map the captured finger motion to control the BrainCo Revo 2 hands.
A GoPro mounted on the G1's head provides egocentric RGB, while the G1's computer records synchronized whole-body expert demonstrations.

\noindent\textbf{Evaluation.}
We evaluate six core SIMPLE tasks~\citep{wei2026simple} at Levels~0--2 with ten rollouts per task--level pair, and four real-world tasks shown in Fig.~\ref{fig:real-world-task-settings}(b), with ten frozen initial configurations per task and policy.
These fixed configurations support consistent comparisons across policies by controlling variation in the initial conditions.
We report task success rate and task progress, capturing both full task completion and partial execution of multi-stage behaviors.
Progress is computed as the sum of fixed milestone scores, normalized to 100 per task and averaged equally across tasks.
Standard trials use robot-seen objects in Fig.~\ref{fig:real-world-task-settings}(c).
Generalization uses robot-unseen objects appearing only in human demonstrations.

\noindent\textbf{Baselines.}
Simulation baselines come from the public SIMPLE leaderboard~\citep{wei2026simple}.
Public entries retain their source protocols. We evaluate StarVLA and \modelname{} ourselves.
Real-world baselines include $\pi_{0.5}$~\citep{physicalintelligence2025pi05}, GR00T N1.6~\citep{gear2025grootn16}, $\Psi_0$~\citep{wei2026psi0}, and StarVLA~\citep{starvlacommunity2026starvla}.
All methods use identical initial configurations and scoring criteria for consistent comparisons.

\vspace{-8pt}
\subsection{Q1: How Well Does the Model Perform on Humanoid Loco-Manipulation?}
\label{sec:exp-performance}

\paragraph{Real-world evaluation.}

Table~\ref{tab:real-robot-results} reports the highest average success rate (62.5\%) and weighted progress (80.4\%) for \modelname{}, exceeding the strongest external baselines by 22.5 and 25.8 percentage points, respectively.
It achieves the highest success count on all four tasks, indicating consistent gains across tasks, with the largest progress improvement on bottle disposal (51.0 points).
Progress remains above 75\% on every task, suggesting broad milestone coverage even where full completion remains challenging.
On chair placement, it completes 6/10 trials versus StarVLA's 3/10 despite lower progress (77.0\% versus 82.5\%), illustrating that partial-task progress and full completion capture different aspects of performance and should be reported together for a fuller assessment.

\begingroup
\setlength{\intextsep}{2pt}
\setlength{\textfloatsep}{6pt}
\begin{table}[!h]
  \centering
  \caption{\textbf{Real-world task performance.}
  We report success rates over ten trials and weighted progress (\%) for each task.
  Averages weight the four tasks equally.}
  \label{tab:real-robot-results}
  \fontsize{7.7}{9.2}\selectfont
  \setlength{\tabcolsep}{1pt}
  \renewcommand{\arraystretch}{1.15}
    \begin{tabular*}{\linewidth}{@{\extracolsep{\fill}}l
      *{3}{c@{\hspace{1pt}}c}!{\hspace{3pt}}
      *{2}{c@{\hspace{1pt}}c}@{}}
      \toprule
      {\scriptsize\textbf{Method}}
        & \multicolumn{2}{c}{\makebox[0pt]{\scriptsize\textbf{Bottle Disposal}}}
        & \multicolumn{2}{c}{\makebox[0pt]{\scriptsize\textbf{Toy Storage}}}
        & \multicolumn{2}{c}{\makebox[0pt]{\scriptsize\textbf{Box Transport}}}
        & \multicolumn{2}{c}{\makebox[0pt]{\scriptsize\textbf{Chair Placement}}}
        & \multicolumn{2}{c}{\makebox[0pt]{\scriptsize\textbf{Average}}} \\
      \midrule
      $\pi_{0.5}$~\citep{physicalintelligence2025pi05}
        & 2/10 & 28.0 & 5/10 & 64.5 & 5/10 & 58.0
        & 4/10 & 65.5 & 16/40 & 54.0 \\
      GR00T N1.6~\citep{gear2025grootn16}
        & 0/10 & 10.0 & 1/10 & 19.0 & 0/10 & 10.0
        & 2/10 & 46.0 & 3/40 & 21.3 \\
      $\Psi_0$~\citep{wei2026psi0}
        & 1/10 & 26.5 & 2/10 & 30.0 & 1/10 & 37.5
        & 0/10 & 33.0 & 4/40 & 31.8 \\
      StarVLA~\citep{starvlacommunity2026starvla}
        & 0/10 & 10.0 & 5/10 & 55.0 & 3/10 & 71.0
        & 3/10 & \textbf{82.5} & 11/40 & 54.6 \\
      \midrule
      $\lambda_0$ \textbf{(Ours)}
        & \textbf{5/10} & \textbf{79.0} & \textbf{7/10} & \textbf{81.0}
        & \textbf{7/10} & \textbf{84.5} & \textbf{6/10} & 77.0
        & \textbf{25/40} & \textbf{80.4} \\
      \bottomrule
    \end{tabular*}
\end{table}

\paragraph{Simulation evaluation.}
\begin{table}
  \centering
  \caption{\textbf{SIMPLE task performance.}
  Success rates (\%) at Levels 0--2 with ten trials per task--level pair.
  Average is computed per level.
  Overall averages all 18 task--level pairs.}
  \label{tab:simple-summary}
  \setlength{\tabcolsep}{3pt}
  \resizebox{\linewidth}{!}{%
    \begin{tabular}{lcccccccc}
      \toprule
      Model & XMovePick & BendPick
      & Handover & Mobile P\&P
      & Grasp & XMoveBendPick
      & Average & Overall \\
      \midrule
      H-RDT~\citep{bi2025hrdt} & 0/0/20 & 0/0/10 & 0/10/0 & 0/0/0 & 0/0/0 & 0/0/0 & 0.0/1.7/5.0 & 2.2 \\
      InternVLA-M1~\citep{chen2025internvlam1} & 0/0/0 & 50/50/0 & 0/0/0 & 0/0/0 & 0/0/0 & 30/50/70 & 13.3/16.7/11.7 & 13.9 \\
      EgoVLA~\citep{yang2025egovla} & 0/10/20 & 70/50/80 & 0/40/30 & 0/0/0 & 100/100/70 & 30/50/40 & 33.3/41.7/40.0 & 38.3 \\
      Diffusion Policy~\citep{chi2025diffusion} & 30/30/20 & 100/80/60 & 30/20/40 & 40/0/0 & 80/90/80 & 0/0/0 & 46.7/36.7/33.3 & 38.9 \\
      GR00T N1.6~\citep{gear2025grootn16} & 100/100/70 & 70/70/60 & 10/30/30 & 0/0/0 & 90/90/70 & 40/40/10 & 51.7/55.0/40.0 & 48.9 \\
      $\pi_{0.5}$~\citep{physicalintelligence2025pi05} & 70/50/10 & 100/100/80 & 50/40/50 & 30/30/30 & 100/100/80 & 0/0/0 & 58.3/53.3/41.7 & 51.1 \\
      DreamZero~\citep{ye2026dreamzero} & 100/100/100 & 90/90/80 & 70/80/90 & 50/30/30 & 90/100/70 & 0/0/10 & 66.7/66.7/63.3 & 65.6 \\
      ACT~\citep{zhao2023aloha} & 100/100/50 & 100/90/90 & 70/70/100 & 50/50/50 & 100/100/80 & 90/100/100 & 85.0/85.0/78.3 & 82.8 \\
      $\Psi_0$~\citep{wei2026psi0} & 100/100/60 & 100/100/100 & 70/70/100 & 70/50/60 & 100/100/80 & 100/90/90 & \textbf{90.0}/85.0/81.7 & 85.6 \\
      StarVLA~\citep{starvlacommunity2026starvla} & 100/90/90 & 70/90/70 & 80/80/100 & 90/80/80 & 90/100/70 & 70/100/100 & 83.3/90.0/85.0 & 86.1 \\
      \midrule
      \rowcolor{black!4}
      \textbf{\modelname{} (Ours)} & 100/100/80 & 100/100/100 & 90/100/100 & 90/70/70 & 100/90/80 & 60/100/90 & \textbf{90.0}/\textbf{93.3}/\textbf{86.7} & \textbf{90.0} \\
      \bottomrule
    \end{tabular}%
  }
\end{table}

Table~\ref{tab:simple-summary} reports 90.0\% overall success for \modelname{}, exceeding StarVLA and $\Psi_0$ by 3.9 and 4.4 percentage points, respectively.
Its success rates of 90.0\%, 93.3\%, and 86.7\% at Levels~0--2 lead StarVLA at every level, showing that the advantage extends across evaluation levels.
The margin over StarVLA narrows to 1.7 points at Level~2, indicating that the relative gain varies with the evaluation setting.
It achieves 100\% success on BendPick at all levels and Handover at Levels~1 and 2, while Mobile P\&P remains below StarVLA at Levels~1 and 2 (70\% versus 80\%), indicating room for improvement in mobile manipulation.
BendPick and Handover maintain strong performance across all three levels.

\vspace{-0.8ex}
\subsection{Q2: How Do Human-Data Scale and the Training Recipe Affect Learning?}
\label{sec:exp-learning}

\paragraph{Validation-loss scaling.}

To isolate human-data scale in Stage II, we train the same 2B model for 100k updates on nested 5\%, 10\%, 25\%, 50\%, and 100\% subsets with a fixed seed and optimization schedule.
Every 1k updates, we evaluate two fixed, session-disjoint holdouts: \emph{clean} contains sessions with upper-quartile whole-body and bilateral-hand annotation validity, while \emph{representative} is stratified by data source and annotation validity to reflect the full data distribution.
Following EgoScale~\citep{zheng2026egoscale}, we analyze training dynamics and minimum loss at each scale.

\setlength{\intextsep}{10pt}
\begin{figure}[!htbp]
  \centering
  \includegraphics[width=0.95\linewidth,trim=0 7bp 0 13bp,clip]{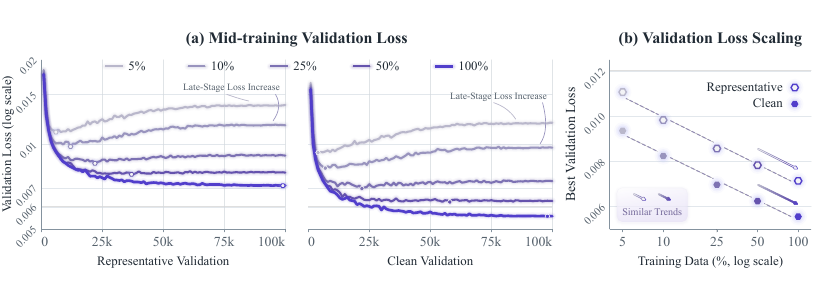}\par
  \setlength{\abovecaptionskip}{2pt}
  \caption{\textbf{Human-data scaling of validation loss.}
  (a) Training curves on representative and clean holdouts at five data fractions.
  (b) Minimum validation loss versus data scale with descriptive log-linear fits over the measured range.
  Minima are selected independently on each holdout.}
  \label{fig:scaling-curves}
\end{figure}
\setlength{\intextsep}{2pt}

\vspace{-6pt}
\paragraph{Training dynamics.}
Fig.~\ref{fig:scaling-curves}(a) shows later validation-loss minima at larger data scales.
At 5\% and 10\%, representative loss reaches its minimum at 6k and 12k updates, then finishes 24.7\% and 19.0\% higher.
On clean, the corresponding increases are 27.6\% and 18.1\%.
At 50\% and 100\%, final losses remain within 1.4\% of the minima on representative and 0.9\% on clean.
At 50\%, the minima occur at 37k/58k updates on representative/clean. At 100\%, they occur at 99k/98k.
The 25\% run forms an intermediate regime: both holdouts reach their minima near 22k updates and finish only 6.4--6.5\% higher.
Larger datasets thus support longer training before validation loss rises, while smaller subsets benefit from earlier checkpoint selection.
Comparing only final checkpoints would confound the effect of data scale with late-stage degradation.

\vspace{-6pt}
\paragraph{Scaling trend.}
Fig.~\ref{fig:scaling-curves}(b) compares minimum validation losses across data scales.
From 5\% to 100\%, loss falls by 35.5\% on representative and 40.6\% on clean.
Fitting $\widehat{L}_h(D)=a_h+b_h\ln D$ gives $R^2=0.9868/0.9901$ and per-doubling reductions of $0.000900/0.000877$ for representative/clean.
Clean has lower loss throughout, while slope magnitudes differ by only 2.7\%, indicating similar scaling responses despite their different annotation-validity distributions.
These single-run fits describe the measured range without establishing cross-seed reproducibility or extrapolation.

\paragraph{Training-stage ablations under fixed robot supervision.}
Table~\ref{tab:real-robot-ablation} ablates the two human-data stages. SIMPLE success is 90.0\% with the full recipe, 87.8\% without Stage I, 82.3\% without Stage II, and 79.4\% without both.
Real-world success/progress is 62.5\%/80.4\% for the full recipe, 50.0\%/75.9\% without Stage I, 22.5\%/59.9\% without Stage II, and 15.0\%/54.0\% without both.
Removing Stage II reduces real-world success more than removing Stage I (40.0 versus 12.5 percentage points), highlighting the importance of whole-body mid-training beyond the interaction priors learned from egocentric hand data.

\begin{table}[!htbp]
  \centering
  \caption{\textbf{Training-stage ablations with fixed robot supervision.}
  Filled/open circles indicate included/removed data. Metrics report SIMPLE success and real-world averages across four tasks.}
  \label{tab:real-robot-ablation}
  \small
  \setlength{\tabcolsep}{2.2pt}
  \resizebox{\linewidth}{!}{%
    \begin{tabular*}{\linewidth}{@{\extracolsep{\fill}}lccccccccc@{}}
      \toprule
      \multirow{2}{*}{Variant}
        & \multirow{2}{*}{S1}
        & \multirow{2}{*}{S2}
        & \multirow{2}{*}{Robot}
        & \multicolumn{4}{c}{SIMPLE (\%)}
        & \multicolumn{2}{c}{Real-world (\%)} \\
      \cmidrule(lr){5-8}\cmidrule(lr){9-10}
        & & & & L0 & L1 & L2 & Overall & Success & Progress \\
      \midrule
      w/o human-data pretraining
        & $\circ$ & $\circ$ & $\bullet$ & 83.3 & 83.3 & 71.7 & 79.4 & 15.0 & 54.0 \\
      w/o Stage II mid-training
        & $\bullet$ & $\circ$ & $\bullet$ & 83.6 & 86.7 & 76.7 & 82.3 & 22.5 & 59.9 \\
      w/o Stage I pretraining
        & $\circ$ & $\bullet$ & $\bullet$ & 88.0 & 91.7 & 83.7 & 87.8 & 50.0 & 75.9 \\
      $\lambda_0$ \textbf{(Full Recipe)}
        & $\bullet$ & $\bullet$ & $\bullet$ & \textbf{90.0} & \textbf{93.3} & \textbf{86.7} & \textbf{90.0} & \textbf{62.5} & \textbf{80.4} \\
      \bottomrule
    \end{tabular*}%
  }
\end{table}

\FloatBarrier

\endgroup
\subsection{Q3: Does Human-Data Scaling Improve Downstream Transfer?}
\label{sec:exp-transfer}

\noindent\begin{minipage}[t]{0.53\linewidth}
\vspace{0pt}
\noindent\textbf{Downstream task-progress scaling.}
Fig.~\ref{fig:downstream-scaling} reports task progress of 59.9\%, 64.2\%, 67.3\%, 72.1\%, 76.4\%, and 80.4\% at 0\%, 5\%, 10\%, 25\%, 50\%, and 100\% Stage-II data, respectively.
The endpoints match the no-Stage-II and full-recipe results in Table~\ref{tab:real-robot-ablation}.
We fix Stage-I initialization, model capacity, robot demonstrations, Stage-III update budget, and evaluation configurations.
Nonzero fractions share Stage-II update budgets and checkpoint selection.
Progress increases monotonically across these fractions, gaining 20.5 percentage points from no Stage II to the full corpus.
\end{minipage}\hfill
\begin{minipage}[t]{0.43\linewidth}
\vspace{0pt}
\begingroup
  \setlength{\parskip}{0pt}
  \centering
  \begin{tikzpicture}
    \begin{axis}[
      width=\linewidth,
      height=4.1cm,
      xmin=0, xmax=5,
      ymin=59, ymax=90,
      xtick={0,1,2,3,4,5},
      xticklabels={0\%,5\%,10\%,25\%,50\%,100\%},
      ytick={60,70,80,90},
      xlabel={Stage-II data fraction},
      ylabel={Task progress (\%)},
      tick label style={font=\scriptsize},
      label style={font=\scriptsize},
      axis line style={black!55},
      tick style={black!55},
      tick align=outside,
      grid=major,
      grid style={black!10},
      clip=false,
    ]
      \addplot[color=referencepurple, thick, mark=*, mark size=2pt]
        coordinates {(0,59.9) (1,64.2) (2,67.3) (3,72.1) (4,76.4) (5,80.4)};
      \node[font=\scriptsize, text=referencepurple, anchor=south west]
        at (axis cs:0,59.9) {59.9};
      \node[font=\scriptsize, text=referencepurple, anchor=south, yshift=2pt]
        at (axis cs:1,64.2) {64.2};
      \node[font=\scriptsize, text=referencepurple, anchor=south, yshift=2pt]
        at (axis cs:2,67.3) {67.3};
      \node[font=\scriptsize, text=referencepurple, anchor=south, yshift=2pt]
        at (axis cs:3,72.1) {72.1};
      \node[font=\scriptsize, text=referencepurple, anchor=south, yshift=2pt]
        at (axis cs:4,76.4) {76.4};
      \node[font=\scriptsize, text=referencepurple, anchor=south east]
        at (axis cs:5,80.4) {80.4};
    \end{axis}
  \end{tikzpicture}\par
  \vspace{-4pt}
  \setlength{\abovecaptionskip}{0pt}
  
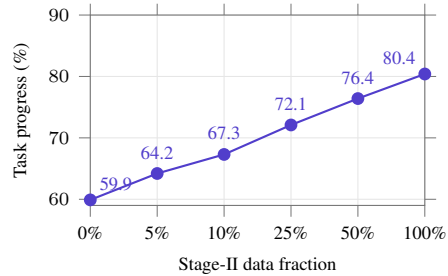
\captionof{figure}{Average real-world task progress versus Stage-II human-data fraction.}
  \label{fig:downstream-scaling}
\endgroup

\end{minipage}\par

\paragraph{Object generalization.}
We evaluate whether human-data training improves transfer to objects absent from robot demonstrations across the four real-world tasks.
For each method, we train a policy using robot demonstrations of objects A and human demonstrations of A and B.
\textit{Seen}~(\transferinline{1}) and \textit{Human-guided}~(\transferinline{2}) evaluate this same policy on A and B, respectively.
\textit{Zero-shot}~(\transferinline{3}) evaluates a robot-only variant on B under matched initializations and scoring.

\begingroup
\setlength{\intextsep}{8pt}
\begin{figure}[!ht]
  \centering
  \setlength{\abovecaptionskip}{0pt}
  \input{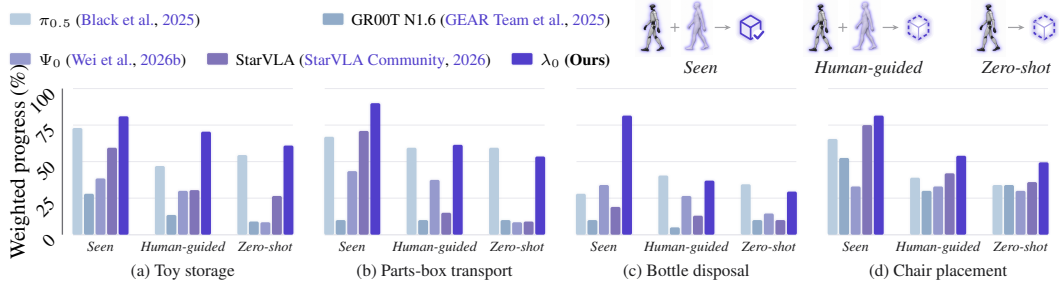}
  \caption{\textbf{Object-transfer progress.}
  Seen and Human-guided evaluate a joint policy trained on robot-seen object A and human-only data for object B.
  Zero-shot tests a robot-only policy on B.}
  \label{fig:generalization-results}
\end{figure}

\endgroup

As shown in Fig.~\ref{fig:generalization-results}, \modelname achieves the highest mean \textit{Human-guided} progress at 55.8\%, compared with 46.5\% for the strongest baseline, $\pi_{0.5}$.
Relative to its \textit{Zero-shot} variant, \modelname improves progress on all four tasks, increasing the average from 48.4\% to 55.8\%, a gain of 7.4 percentage points.
These results support human-data training as a means of transferring manipulation skills to robot-unseen objects, although performance remains below the \textit{Seen} setting.

\section{Conclusion}
We presented \modelname, a whole-body humanoid vision--language--action policy that transfers human experience through three-stage training.
We introduced \datasetname, a 500-hour dataset of diverse open-world human activities with synchronized egocentric video, body motion, and hand motion.
Experiments on SIMPLE and four real-world tasks demonstrated improved loco-manipulation performance and transfer to robot-unseen objects.
Controlled experiments revealed consistent benefits from scaling whole-body human data and highlighted the contribution of whole-body mid-training.
These findings support scalable human activity as rich supervision for humanoid control.
\paragraph{Limitations and future work.}
Our evaluation focuses on G1, leaving transfer to full-sized humanoids untested.
Despite camera alignment and motion retargeting, differences in body proportions, viewpoints, and feasible motions leave human--robot alignment imperfect.
Future work will extend \modelname to full-sized humanoids through improved motion retargeting and closer alignment between egocentric observations, whole-body trajectories, and physically executable robot actions.

\begingroup
\hypersetup{urlcolor=black,linkcolor=black,citecolor=black}
\bibliography{references}

@inproceedings{dasari2020robonet,
  title        = {{RoboNet}: Large-Scale Multi-Robot Learning},
  author       = {Dasari, Sudeep and Ebert, Frederik and Tian, Stephen and Nair, Suraj and Bucher, Bernadette and Schmeckpeper, Karl and Singh, Siddharth and Levine, Sergey and Finn, Chelsea},
  booktitle    = {Proceedings of the Conference on Robot Learning},
  year         = {2020},
  volume       = {100},
  pages        = {885--897},
}

@inproceedings{jang2022bcz,
  title        = {{BC-Z}: Zero-Shot Task Generalization with Robotic Imitation Learning},
  author       = {Jang, Eric and Irpan, Alex and Khansari, Mohi and Kappler, Daniel and Ebert, Frederik and Lynch, Corey and Levine, Sergey and Finn, Chelsea},
  booktitle    = {Proceedings of the Conference on Robot Learning},
  year         = {2022},
  volume       = {164},
  pages        = {991--1002},
}

@inproceedings{walke2023bridgedata,
  title        = {{BridgeData V2}: A Dataset for Robot Learning at Scale},
  author       = {Walke, Homer Rich and Black, Kevin and Zhao, Tony Z. and Vuong, Quan and Zheng, Chongyi and Hansen-Estruch, Philippe and He, Andre Wang and Myers, Vivek and Kim, Moo Jin and Du, Max and Lee, Abraham and Fang, Kuan and Finn, Chelsea and Levine, Sergey},
  booktitle    = {Proceedings of the Conference on Robot Learning},
  year         = {2023},
  volume       = {229},
  pages        = {1723--1736},
}

@inproceedings{openx2024,
  title        = {Open {X}-Embodiment: Robotic Learning Datasets and {RT-X} Models},
  author       = {{Open X-Embodiment Collaboration}},
  booktitle    = {Proceedings of the IEEE International Conference on Robotics and Automation},
  year         = {2024},
  pages        = {6892--6903},
  doi          = {10.1109/icra57147.2024.10611477},
}

@inproceedings{brohan2023rt1,
  title        = {{RT-1}: Robotics Transformer for Real-World Control at Scale},
  author       = {Brohan, Anthony and others},
  booktitle    = {Proceedings of Robotics: Science and Systems},
  year         = {2023},
  volume       = {19},
  doi          = {10.15607/RSS.2023.XIX.025},
}

@inproceedings{zitkovich2023rt2,
  title        = {{RT-2}: Vision-Language-Action Models Transfer Web Knowledge to Robotic Control},
  author       = {Zitkovich, Brianna and others},
  booktitle    = {Proceedings of the Conference on Robot Learning},
  year         = {2023},
  volume       = {229},
  pages        = {2165--2183},
}

@inproceedings{khazatsky2024droid,
  title        = {{DROID}: A Large-Scale In-the-Wild Robot Manipulation Dataset},
  author       = {Khazatsky, Alexander and others},
  booktitle    = {Proceedings of Robotics: Science and Systems},
  year         = {2024},
  volume       = {20},
  doi          = {10.15607/RSS.2024.XX.120},
}

@inproceedings{bu2025agibot,
  title        = {{AgiBot World Colosseo}: A Large-Scale Manipulation Platform for Scalable and Intelligent Embodied Systems},
  author       = {Bu, Qingwen and others},
  booktitle    = {Proceedings of the IEEE/RSJ International Conference on Intelligent Robots and Systems},
  year         = {2025},
  pages        = {3549--3556},
  doi          = {10.1109/iros60139.2025.11247088},
}

@inproceedings{ghosh2024octo,
  title        = {{Octo}: An Open-Source Generalist Robot Policy},
  author       = {Ghosh, Dibya and Walke, Homer Rich and Pertsch, Karl and Black, Kevin and Mees, Oier and Dasari, Sudeep and Hejna, Joey and Kreiman, Tobias and Xu, Charles and Luo, Jianlan and Tan, You Liang and Chen, Lawrence Yunliang and Vuong, Quan and Xiao, Ted and Sanketi, Pannag R. and Sadigh, Dorsa and Finn, Chelsea and Levine, Sergey},
  booktitle    = {Proceedings of Robotics: Science and Systems},
  year         = {2024},
  volume       = {20},
  doi          = {10.15607/RSS.2024.XX.090},
}

@inproceedings{kim2025openvla,
  title        = {{OpenVLA}: An Open-Source Vision-Language-Action Model},
  author       = {Kim, Moo Jin and others},
  booktitle    = {Proceedings of the Conference on Robot Learning},
  year         = {2025},
  volume       = {270},
  pages        = {2679--2713},
}

@article{chi2025diffusion,
  title        = {Diffusion Policy: Visuomotor Policy Learning via Action Diffusion},
  author       = {Chi, Cheng and Xu, Zhenjia and Feng, Siyuan and Cousineau, Eric and Du, Yilun and Burchfiel, Benjamin and Tedrake, Russ and Song, Shuran},
  journal      = {The International Journal of Robotics Research},
  year         = {2025},
  volume       = {44},
  number       = {10--11},
  pages        = {1684--1704},
  doi          = {10.1177/02783649241273668},
}

@inproceedings{dass2023pato,
  title        = {{PATO}: Policy Assisted TeleOperation for Scalable Robot Data Collection},
  author       = {Dass, Shivin and Pertsch, Karl and Zhang, Hejia and Lee, Youngwoon and Lim, Joseph J. and Nikolaidis, Stefanos},
  booktitle    = {Proceedings of Robotics: Science and Systems},
  year         = {2023},
  volume       = {19},
  doi          = {10.15607/RSS.2023.XIX.013},
}

@inproceedings{qin2023anyteleop,
  title        = {{AnyTeleop}: A General Vision-Based Dexterous Robot Arm-Hand Teleoperation System},
  author       = {Qin, Yuzhe and Yang, Wei and Huang, Binghao and Van Wyk, Karl and Su, Hao and Wang, Xiaolong and Chao, Yu-Wei and Fox, Dieter},
  booktitle    = {Proceedings of Robotics: Science and Systems},
  year         = {2023},
  volume       = {19},
  doi          = {10.15607/RSS.2023.XIX.015},
}

@inproceedings{zhao2023aloha,
  title        = {Learning Fine-Grained Bimanual Manipulation with Low-Cost Hardware},
  author       = {Zhao, Tony Z. and Kumar, Vikash and Levine, Sergey and Finn, Chelsea},
  booktitle    = {Proceedings of Robotics: Science and Systems},
  year         = {2023},
  volume       = {19},
  doi          = {10.15607/RSS.2023.XIX.016},
}

@inproceedings{fu2025mobilealoha,
  title        = {Mobile {ALOHA}: Learning Bimanual Mobile Manipulation using Low-Cost Whole-Body Teleoperation},
  author       = {Fu, Zipeng and Zhao, Tony Z. and Finn, Chelsea},
  booktitle    = {Proceedings of the Conference on Robot Learning},
  year         = {2025},
  volume       = {270},
  pages        = {4066--4083},
}

@inproceedings{chi2024umi,
  title        = {Universal Manipulation Interface: In-the-Wild Robot Teaching Without In-the-Wild Robots},
  author       = {Chi, Cheng and Xu, Zhenjia and Pan, Chuer and Cousineau, Eric and Burchfiel, Benjamin and Feng, Siyuan and Tedrake, Russ and Song, Shuran},
  booktitle    = {Proceedings of Robotics: Science and Systems},
  year         = {2024},
  volume       = {20},
  doi          = {10.15607/RSS.2024.XX.045},
}

@inproceedings{grauman2022ego4d,
  title        = {{Ego4D}: Around the World in 3,000 Hours of Egocentric Video},
  author       = {Grauman, Kristen and others},
  booktitle    = {Proceedings of the IEEE/CVF Conference on Computer Vision and Pattern Recognition},
  year         = {2022},
  pages        = {18995--19012},
}

@article{damen2022epic,
  title        = {Rescaling Egocentric Vision: Collection, Pipeline and Challenges for {EPIC-KITCHENS-100}},
  author       = {Damen, Dima and others},
  journal      = {International Journal of Computer Vision},
  year         = {2022},
  volume       = {130},
  number       = {1},
  pages        = {33--55},
  doi          = {10.1007/s11263-021-01531-2},
}

@inproceedings{goyal2017something,
  title        = {The ``Something Something'' Video Database for Learning and Evaluating Visual Common Sense},
  author       = {Goyal, Raghav and others},
  booktitle    = {Proceedings of the International Conference on Computer Vision},
  year         = {2017},
  pages        = {5842--5850},
}

@inproceedings{grauman2024egoexo4d,
  title        = {{Ego-Exo4D}: Understanding Skilled Human Activity from First- and Third-Person Perspectives},
  author       = {Grauman, Kristen and others},
  booktitle    = {Proceedings of the IEEE/CVF Conference on Computer Vision and Pattern Recognition},
  year         = {2024},
  pages        = {19383--19400},
}

@inproceedings{wang2023holoassist,
  title        = {{HoloAssist}: An Egocentric Human Interaction Dataset for Interactive {AI} Assistants in the Real World},
  author       = {Wang, Xin and others},
  booktitle    = {Proceedings of the International Conference on Computer Vision},
  year         = {2023},
  pages        = {20270--20281},
}

@inproceedings{liu2022hoi4d,
  title        = {{HOI4D}: A {4D} Egocentric Dataset for Category-Level Human-Object Interaction},
  author       = {Liu, Yunze and Liu, Yun and Jiang, Che and Lyu, Kangbo and Wan, Weikang and Shen, Hao and Liang, Boqiang and Fu, Zhoujie and Wang, He and Yi, Li},
  booktitle    = {Proceedings of the IEEE/CVF Conference on Computer Vision and Pattern Recognition},
  year         = {2022},
  pages        = {21013--21022},
}

@inproceedings{wang2025hocap,
  title        = {{HO-Cap}: A Capture System and Dataset for {3D} Reconstruction and Pose Tracking of Hand-Object Interaction},
  author       = {Wang, Jikai and Zhang, Qifan and Chao, Yu-Wei and Wen, Bowen and Guo, Xiaohu and Xiang, Yu},
  booktitle    = {Advances in Neural Information Processing Systems},
  year         = {2025},
  volume       = {38},
  doi          = {10.52202/085713-5683},
}

@inproceedings{sener2022assembly101,
  title        = {{Assembly101}: A Large-Scale Multi-View Video Dataset for Understanding Procedural Activities},
  author       = {Sener, Fadime and Chatterjee, Dibyadip and Shelepov, Daniel and He, Kun and Singhania, Dipika and Wang, Robert and Yao, Angela},
  booktitle    = {Proceedings of the IEEE/CVF Conference on Computer Vision and Pattern Recognition},
  year         = {2022},
  pages        = {21096--21106},
}

@inproceedings{ragusa2021meccano,
  title        = {The {MECCANO} Dataset: Understanding Human-Object Interactions from Egocentric Videos in an Industrial-Like Domain},
  author       = {Ragusa, Francesco and Furnari, Antonino and Livatino, Salvatore and Farinella, Giovanni Maria},
  booktitle    = {Proceedings of the IEEE/CVF Winter Conference on Applications of Computer Vision},
  year         = {2021},
  pages        = {1569--1578},
}

@inproceedings{huang2024egoexolearn,
  title        = {{EgoExoLearn}: A Dataset for Bridging Asynchronous Ego- and Exo-centric View of Procedural Activities in Real World},
  author       = {Huang, Yifei and others},
  booktitle    = {Proceedings of the IEEE/CVF Conference on Computer Vision and Pattern Recognition},
  year         = {2024},
  pages        = {22072--22086},
}

@inproceedings{banerjee2025hot3d,
  title        = {{HOT3D}: Hand and Object Tracking in {3D} from Egocentric Multi-View Videos},
  author       = {Banerjee, Prithviraj and others},
  booktitle    = {Proceedings of the IEEE/CVF Conference on Computer Vision and Pattern Recognition},
  year         = {2025},
  pages        = {7061--7071},
}

@inproceedings{hoque2025egodex,
  title        = {{EgoDex}: Learning Dexterous Manipulation from Large-Scale Egocentric Video},
  author       = {Hoque, Ryan and Huang, Peide and Yoon, David J. and Sivapurapu, Mouli and Zhang, Jian},
  booktitle    = {Proceedings of the International Conference on Learning Representations},
  year         = {2026},
}

@inproceedings{punamiya2026egoverse,
  title        = {{EgoVerse}: An Egocentric Human Dataset for Robot Learning from Around the World},
  author       = {Punamiya, Ryan and others},
  booktitle    = {Proceedings of Robotics: Science and Systems},
  year         = {2026},
  volume       = {22},
  doi          = {10.15607/RSS.2026.XXII.092},
}

@inproceedings{li2025vitra,
  title        = {Scalable Vision-Language-Action Model Pretraining for Robotic Manipulation with Real-Life Human Activity Videos},
  author       = {Li, Qixiu and others},
  booktitle    = {Proceedings of the IEEE International Conference on Robotics and Automation},
  year         = {2026},
}

@inproceedings{zheng2026egoscale,
  title        = {{EgoScale}: Scaling Dexterous Manipulation with Diverse Egocentric Human Data},
  author       = {Zheng, Ruijie and others},
  booktitle    = {Proceedings of the Conference on Robot Learning},
  year         = {2026},
  note         = {To appear},
}

@article{wang2026ego2robot,
  title        = {{Ego2Robot}: Scalable Robot Data Synthesis from Egocentric Human Data},
  author       = {Wang, Ye and Lin, Pei and Chen, Xiong-Hui and others},
  journal      = {arXiv preprint arXiv:2608.02580},
  year         = {2026},
}

@inproceedings{nair2023r3m,
  title        = {{R3M}: A Universal Visual Representation for Robot Manipulation},
  author       = {Nair, Suraj and Rajeswaran, Aravind and Kumar, Vikash and Finn, Chelsea and Gupta, Abhinav},
  booktitle    = {Proceedings of the Conference on Robot Learning},
  year         = {2023},
  volume       = {205},
  pages        = {892--909},
}

@inproceedings{sontakke2023roboclip,
  title        = {{RoboCLIP}: One Demonstration Is Enough to Learn Robot Policies},
  author       = {Sontakke, Sumedh and Zhang, Jesse and Arnold, S{\'e}b and Pertsch, Karl and B{\i}y{\i}k, Erdem and Sadigh, Dorsa and Finn, Chelsea and Itti, Laurent},
  booktitle    = {Advances in Neural Information Processing Systems},
  year         = {2023},
  volume       = {36},
  pages        = {55681--55693},
  doi          = {10.52202/075280-2430},
}

@inproceedings{kumar2020subroutines,
  title        = {Learning Navigation Subroutines from Egocentric Videos},
  author       = {Kumar, Ashish and Gupta, Saurabh and Malik, Jitendra},
  booktitle    = {Proceedings of the Conference on Robot Learning},
  year         = {2020},
  volume       = {100},
  pages        = {617--626},
}

@inproceedings{shao2020concept2robot,
  title        = {{Concept2Robot}: Learning Manipulation Concepts from Instructions and Human Demonstrations},
  author       = {Shao, Lin and Migimatsu, Toki and Zhang, Qiang and Yang, Kaiyuan and Bohg, Jeannette},
  booktitle    = {Proceedings of Robotics: Science and Systems},
  year         = {2020},
  volume       = {16},
  doi          = {10.15607/RSS.2020.XVI.082},
}

@inproceedings{bahl2022whirl,
  title        = {Human-to-Robot Imitation in the Wild},
  author       = {Bahl, Shikhar and Gupta, Abhinav and Pathak, Deepak},
  booktitle    = {Proceedings of Robotics: Science and Systems},
  year         = {2022},
  volume       = {18},
  doi          = {10.15607/RSS.2022.XVIII.026},
}

@inproceedings{wang2023mimicplay,
  title        = {{MimicPlay}: Long-Horizon Imitation Learning by Watching Human Play},
  author       = {Wang, Chen and Fan, Linxi and Sun, Jiankai and Zhang, Ruohan and Fei-Fei, Li and Xu, Danfei and Zhu, Yuke and Anandkumar, Anima},
  booktitle    = {Proceedings of the Conference on Robot Learning},
  year         = {2023},
  volume       = {229},
  pages        = {201--221},
}

@article{bharadhwaj2023zeroshot,
  title        = {Zero-Shot Robot Manipulation from Passive Human Videos},
  author       = {Bharadhwaj, Homanga and Gupta, Abhinav and Tulsiani, Shubham and Kumar, Vikash},
  journal      = {arXiv preprint arXiv:2302.02011},
  year         = {2023},
}

@inproceedings{kareer2024egomimic,
  title        = {{EgoMimic}: Scaling Imitation Learning via Egocentric Video},
  author       = {Kareer, Simar and Patel, Dhruv and Punamiya, Ryan and Mathur, Pranay and Cheng, Shuo and Wang, Chen and Hoffman, Judy and Xu, Danfei},
  booktitle    = {Proceedings of the IEEE International Conference on Robotics and Automation},
  year         = {2025},
  pages        = {13226--13233},
  doi          = {10.1109/icra55743.2025.11127989},
}

@inproceedings{lepert2025phantom,
  title        = {Phantom: Training Robots Without Robots Using Only Human Videos},
  author       = {Lepert, Marion and Fang, Jiaying and Bohg, Jeannette},
  booktitle    = {Proceedings of the Conference on Robot Learning},
  year         = {2025},
  volume       = {305},
  pages        = {4545--4565},
}

@inproceedings{mandlekar2022offline,
  title        = {What Matters in Learning from Offline Human Demonstrations for Robot Manipulation},
  author       = {Mandlekar, Ajay and Xu, Danfei and Wong, Josiah and Nasiriany, Soroush and Wang, Chen and Kulkarni, Rohun and Fei-Fei, Li and Savarese, Silvio and Zhu, Yuke and Mart{\'i}n-Mart{\'i}n, Roberto},
  booktitle    = {Proceedings of the Conference on Robot Learning},
  year         = {2022},
  volume       = {164},
  pages        = {1678--1690},
}

@inproceedings{fu2025humanplus,
  title        = {{HumanPlus}: Humanoid Shadowing and Imitation from Humans},
  author       = {Fu, Zipeng and Zhao, Qingqing and Wu, Qi and Wetzstein, Gordon and Finn, Chelsea},
  booktitle    = {Proceedings of the Conference on Robot Learning},
  year         = {2025},
  volume       = {270},
  pages        = {2828--2844},
}

@inproceedings{he2025omnih2o,
  title        = {{OmniH2O}: Universal and Dexterous Human-to-Humanoid Whole-Body Teleoperation and Learning},
  author       = {He, Tairan and Luo, Zhengyi and He, Xialin and Xiao, Wenli and Zhang, Chong and Zhang, Weinan and Kitani, Kris M. and Liu, Changliu and Shi, Guanya},
  booktitle    = {Proceedings of the Conference on Robot Learning},
  year         = {2025},
  volume       = {270},
  pages        = {1516--1540},
}

@inproceedings{shi2026egohumanoid,
  title        = {Unlocking In-the-Wild Loco-Manipulation with Robot-Free Egocentric Demonstration},
  author       = {Shi, Modi and others},
  booktitle    = {Proceedings of Robotics: Science and Systems},
  year         = {2026},
  volume       = {22},
  doi          = {10.15607/RSS.2026.XXII.204},
}

@inproceedings{jiang2026wholebodyvla,
  title        = {{WholeBodyVLA}: Towards Unified Latent {VLA} for Whole-Body Loco-Manipulation Control},
  author       = {Jiang, Haoran and Chen, Jin and Bu, Qingwen and Chen, Li and Shi, Modi and Zhang, Yanjie and Li, Delong and Suo, Chuanzhe and Wang, Chuang and Peng, Zhihui and Li, Hongyang},
  booktitle    = {Proceedings of the International Conference on Learning Representations},
  year         = {2026},
}

@article{bjorck2025groot,
  title        = {{GR00T N1}: An Open Foundation Model for Generalist Humanoid Robots},
  author       = {Bjorck, Johan and others},
  journal      = {arXiv preprint arXiv:2503.14734},
  year         = {2025},
}

@inproceedings{wei2026psi0,
  title        = {{$\Psi_0$}: An Open Foundation Model Towards Universal Humanoid Loco-Manipulation},
  author       = {Wei, Songlin and others},
  booktitle    = {Proceedings of Robotics: Science and Systems},
  year         = {2026},
  volume       = {22},
  doi          = {10.15607/RSS.2026.XXII.021},
}

@article{hu2026openhlm,
  title        = {{OpenHLM}: An Empirical Recipe for Whole-Body Humanoid Loco-Manipulation},
  author       = {Hu, Yingdong and others},
  journal      = {arXiv preprint arXiv:2606.22174},
  year         = {2026},
}

@article{li2026omega0,
  title        = {{$\omega$}-0: A Latent Predictive World Action Model for Concurrent Humanoid Loco-Manipulation},
  author       = {Li, Zhe and Zhang, Zhenzhe and Wei, Yangyang and Zhang, Wenjie and Yuan, Xichen and Zhi, Peiyuan and Li, Gen and Guo, Xinying and Gao, Fengjie and Yang, Jianfei and Zhang, Shanghang},
  journal      = {arXiv preprint arXiv:2608.06375},
  year         = {2026},
}

@article{luo2025sonic,
  title        = {{SONIC}: Supersizing Motion Tracking for Natural Humanoid Whole-Body Control},
  author       = {Luo, Zhengyi and others},
  journal      = {Science Robotics},
  year         = {2026},
  volume       = {11},
  number       = {117},
  pages        = {eaed4592},
  doi          = {10.1126/scirobotics.aed4592},
}

@inproceedings{wei2026simple,
  title        = {{SIMPLE}: Simulation-Based Policy Learning and Evaluation for Humanoid Loco-Manipulation},
  author       = {Wei, Songlin and Ni, Zhenhao and Liu, Jie and Zhao, Zhenyu and Ye, Junjie and Jing, Hongyi and Xia, Junkai and Liu, Xiawei and Leong, Michael and Heng, Liang and Huang, Di and Wang, Yue},
  booktitle    = {Proceedings of the Conference on Robot Learning},
  year         = {2026},
  note         = {To appear},
}

@misc{gear2025grootn16,
  title        = {{GR00T N1.6}: An Improved Open Foundation Model for Generalist Humanoid Robots},
  author       = {{GEAR Team} and Azzolini, Allison and Bjorck, Johan and others},
  howpublished = {NVIDIA Research},
  year         = {2025},
  url          = {https://research.nvidia.com/labs/gear/gr00t-n1_6/},
  note         = {Accessed September 26, 2026},
}

@inproceedings{physicalintelligence2025pi05,
  title        = {{$\pi_{0.5}$}: A Vision-Language-Action Model with Open-World Generalization},
  author       = {Black, Kevin and Brown, Noah and Darpinian, James and Dhabalia, Karan and Driess, Danny and Esmail, Adnan and Equi, Michael Robert and Finn, Chelsea and others},
  booktitle    = {Proceedings of the Conference on Robot Learning},
  year         = {2025},
  volume       = {305},
  pages        = {17--40},
}

@article{chen2025internvlam1,
  title        = {{InternVLA-M1}: A Spatially Guided Vision-Language-Action Framework for Generalist Robot Policy},
  author       = {Chen, Xinyi and Chen, Yilun and Fu, Yanwei and Gao, Ning and Jia, Jiaya and Jin, Weiyang and Li, Hao and Mu, Yao and Pang, Jiangmiao and Qiao, Yu and others},
  journal      = {arXiv preprint arXiv:2510.13778},
  year         = {2025},
}

@inproceedings{bi2025hrdt,
  title        = {{H-RDT}: Human Manipulation Enhanced Bimanual Robotic Manipulation},
  author       = {Bi, Hongzhe and Wu, Lingxuan and Lin, Tianwei and Tan, Hengkai and Su, Zhizhong and Su, Hang and Zhu, Jun},
  booktitle    = {Proceedings of the AAAI Conference on Artificial Intelligence},
  year         = {2026},
  volume       = {40},
  pages        = {18135--18143},
  doi          = {10.1609/aaai.v40i22.38875},
}

@inproceedings{ye2026dreamzero,
  title        = {World Action Models Are Zero-Shot Policies},
  author       = {Ye, Seonghyeon and Ge, Yunhao and Zheng, Kaiyuan and Gao, Shenyuan and Yu, Sihyun and Kurian, George and Indupuru, Suneel and Tan, You Liang and Zhu, Chuning and Xiang, Jiannan and others},
  booktitle    = {Proceedings of the Conference on Robot Learning},
  year         = {2026},
  note         = {To appear},
}

@inproceedings{yang2025egovla,
  title        = {{EgoVLA}: Learning Vision-Language-Action Models from Egocentric Human Videos},
  author       = {Yang, Ruihan and Yu, Qinxi and Wu, Yecheng and Yan, Rui and Li, Borui and Cheng, An-Chieh and Zou, Xueyan and Fang, Yunhao and Cheng, Xuxin and Qiu, Ri-Zhao and others},
  booktitle    = {Proceedings of the IEEE/RSJ International Conference on Intelligent Robots and Systems},
  year         = {2026},
  note         = {To appear},
}

@article{starvlacommunity2026starvla,
  title        = {{StarVLA}: A Lego-like Codebase for Vision-Language-Action Model Developing},
  author       = {{StarVLA Community}},
  journal      = {arXiv preprint arXiv:2604.05014},
  year         = {2026},
}

@misc{symbiosis2026dpc,
  title        = {Direct Perception Control Model},
  author       = {{Symbiosis Robotics}},
  howpublished = {Symbiosis Robotics},
  year         = {2026},
  url          = {https://symbiosis-robotics.com/research/dpc/en/},
  note         = {Accessed September 26, 2026},
}

@misc{figure2026helix02,
  title        = {Introducing {Helix 02}: Full-Body Autonomy},
  author       = {{Figure AI}},
  howpublished = {Figure AI},
  year         = {2026},
  month        = jan,
  url          = {https://www.figure.ai/news/helix-02},
  note         = {Accessed September 26, 2026},
}

@misc{parada2026geminirobotics2,
  title        = {{Gemini Robotics 2} Brings Whole Body Intelligence to Robots},
  author       = {Parada, Carolina},
  howpublished = {Google DeepMind},
  year         = {2026},
  month        = jul,
  url          = {https://deepmind.google/blog/gemini-robotics-2-brings-whole-body-intelligence-to-robots/},
  note         = {Accessed September 26, 2026},
}

@article{detone2026nymeriaplus,
  title        = {{NymeriaPlus}: Enriching Nymeria Dataset with Additional Annotations and Data},
  author       = {DeTone, Daniel and Bogo, Federica and Le, Eric-Tuan and Frost, Duncan and Straub, Julian and Siddiqui, Yawar and Ye, Yuting and Engel, Jakob and Newcombe, Richard and Ma, Lingni},
  journal      = {arXiv preprint arXiv:2603.18496},
  year         = {2026},
}

@misc{figure2026index,
  title        = {Introducing {Index}: Building the World's Largest and Most Diverse Physical Dataset},
  author       = {{Figure}},
  howpublished = {Figure AI},
  year         = {2026},
  month        = aug,
  url          = {https://www.figure.ai/news/introducing-index},
  note         = {Accessed September 26, 2026},
}

@misc{lightwheel2026egosuite,
  title        = {{EgoSuite-Open100K}: 100,000 Hours of Egocentric Human Data for Physical {AI}},
  author       = {{Lightwheel AI}},
  howpublished = {Lightwheel AI},
  year         = {2026},
  month        = aug,
  url          = {https://huggingface.co/blog/LightwheelAI/egosuite-open100k},
  note         = {Accessed September 26, 2026},
}

@article{zheng2026motionwam,
  title        = {{MotionWAM}: Towards Foundation World Action Models for Real-Time Humanoid Loco-Manipulation},
  author       = {Zheng, Jia and Ma, Teli and Fan, Yudong and Wang, Zifan and Yang, Shuo and Liang, Junwei},
  journal      = {arXiv preprint arXiv:2606.09215},
  year         = {2026},
}

@inproceedings{zhao2025humanoideveryday,
  title        = {{Humanoid Everyday}: A Comprehensive Robotic Dataset for Open-World Humanoid Manipulation},
  author       = {Zhao, Zhenyu and others},
  booktitle    = {Proceedings of the IEEE International Conference on Robotics and Automation},
  year         = {2026},
}

@inproceedings{beingbeyond2026beingh,
  title        = {{Being-H0.5}: Scaling Human-Centric Robot Learning for Cross-Embodiment Generalization},
  author       = {Luo, Hao and Wang, Ye and Zhang, Wanpeng and Zheng, Sipeng and Xi, Ziheng and Xu, Chaoyi and Xu, Haiweng and Yuan, Haoqi and Zhang, Chi and Wang, Yiqing and Feng, Yicheng and Lu, Zongqing},
  booktitle    = {Proceedings of the Conference on Robot Learning},
  year         = {2026},
  note         = {To appear},
}

@inproceedings{nvidia2026dreamdojo,
  title        = {{DreamDojo}: A Generalist Robot World Model from Large-Scale Human Videos},
  author       = {Gao, Shenyuan and Liang, William and Zheng, Kaiyuan and others},
  booktitle    = {Proceedings of the International Conference on Machine Learning},
  year         = {2026},
}

@misc{figure2026helix25,
  title        = {{Helix 2.5}: Zero-Shot 30-Home Generalization},
  author       = {{Figure AI}},
  howpublished = {Figure AI},
  year         = {2026},
  month        = sep,
  url          = {https://www.figure.ai/news/helix-2-5-zero-shot-30-home-generalization},
  note         = {Published September 17, 2026. Accessed September 26, 2026},
}

@article{zhao2026halomi,
  title   = {{HALOMI}: Learning Humanoid Loco-Manipulation with Active Perception from Human Demonstrations},
  author  = {Zhao, Zehui and Zhao, Yuxuan and Zhang, Gaojing and Liu, Chenxi and Zheng, Maolin and Lian, Wenzhao},
  journal = {arXiv preprint arXiv:2606.18772},
  year    = {2026},
  url     = {https://arxiv.org/abs/2606.18772},
}

@misc{pico_body_tracking,
  title        = {Body Tracking},
  author       = {{PICO}},
  howpublished = {PICO Developer Documentation},
  year         = {n.d.},
  url          = {https://developer.picoxr.com/document/unity/body-tracking/},
  note         = {Accessed September 26, 2026},
}

@inproceedings{yu2025dynhamr,
  title        = {{Dyn-HaMR}: Recovering {4D} Interacting Hand Motion from a Dynamic Camera},
  author       = {Yu, Zhengdi and Zafeiriou, Stefanos and Birdal, Tolga},
  booktitle    = {Proceedings of the IEEE/CVF Conference on Computer Vision and Pattern Recognition},
  month        = jun,
  year         = {2025},
  url          = {https://dyn-hamr.github.io/},
}

@inproceedings{huang2025vipe,
  title        = {{ViPE}: Video Pose Engine for {3D} Geometric Perception},
  author       = {Huang, Jiahui and Zhou, Qunjie and Rabeti, Hesam and Korovko, Aleksandr and Ling, Huan and Ren, Xuanchi and Shen, Tianchang and Gao, Jun and Slepichev, Dmitry and Lin, Chen-Hsuan and Ren, Jiawei and Xie, Kevin and Biswas, Joydeep and Leal-Taixe, Laura and Fidler, Sanja},
  booktitle    = {NVIDIA Research Whitepapers},
  year         = {2025},
  url          = {https://research.nvidia.com/labs/toronto-ai/vipe/},
}

@misc{brainco_revo2,
  title        = {{Revo 2}: Overview},
  author       = {{BrainCo}},
  howpublished = {BrainCo Hand Documentation},
  year         = {n.d.},
  url          = {https://www.brainco-hz.com/docs/revolimb-hand-dev/en/revo2/parameters.html},
  note         = {Accessed September 26, 2026},
}

@misc{qwen2026qwen35,
  title  = {{Qwen3.5}: Towards Native Multimodal Agents},
  author = {{Qwen Team}},
  month  = feb,
  year   = {2026},
  url    = {https://qwen.ai/blog?id=qwen3.5},
}
\bibliographystyle{iclr2027_conference}
\endgroup

\end{document}